\documentclass[preprint,12pt]{elsarticle}

\usepackage{amssymb}
\usepackage{amsmath}
\usepackage{multirow}
\usepackage{booktabs}
\usepackage[normalem]{ulem}

\journal{arXiv}

\begin{document}

\begin{frontmatter}



\title{Selective clustering ensemble based on kappa and F-score}


\author[label1]{Jie Yan}

\author[label2]{Xin Liu}

\author[label1]{Ji Qi}

\author[label3]{Tao You}

\author[label1]{Zhong-Yuan Zhang\corref{mycorrespondingauthor}}

\address[label1]{School of Statistics and Mathematics, \\ Central University of Finance and Economics, Beijing, P.R.China}

\address[label2]{School of Computer and Information Technology, \\ Beijing Jiaotong University, Beijing, P.R.China}

\address[label3]{Bosera Asset Management Co., Ltd., Beijing, P.R.China.}

\cortext[mycorrespondingauthor]{Corresponding author}
\ead{zhyuanzh@gmail.com}

\begin{abstract}
Clustering ensemble has an impressive performance in improving the accuracy and robustness of partition results and has received much attention in recent years. Selective clustering ensemble (SCE) can further improve the ensemble performance by selecting base partitions or clusters in according to diversity and stability. However, there is a conflict between diversity and stability, and how to make the trade-off between the two is challenging. The key here is how to evaluate the quality of the base partitions and clusters. In this paper, we propose a new evaluation method for partitions and clusters using kappa and F-score, leading to a new SCE method, which  uses kappa to select informative base partitions and uses F-score to weight clusters based on stability. The effectiveness and efficiency of the proposed method is empirically validated over real datasets.
\end{abstract}

\begin{highlights}
\item
This paper analyzed how to evaluate the quality of the base partitions and clusters in Selective Clustering Ensemble (SCE) and Weighted Clustering Ensemble (WCE).

\item
We proposed a new evaluation method for partitions and clusters using kappa and F-score, leading to a new SCE method, which is called Diversity-Stability-Kappa-F method (DSKF).

\item
The effectiveness and efficiency of the proposed method was empirically validated over real datasets.

\item
Based on the empirical results, we observed that Normalized Mutual Information (NMI) values were misleading and failed to accurately reflect the actual performance of the partitions, and kappa is a better choice.

\end{highlights}


\end{frontmatter}


\section{Introduction}
Clustering is a method of dividing objects into groups with high intra-cluster similarity and low inter-cluster similarity \cite{jain1999data}. As an effective way to understand the underlying structure of a given dataset, it plays a critical role in plentiful fields, such as pattern recognition, information retrieval and recommender systems, etc. Specifically, it is often used for data mining of raw data with few prior knowledge, or as a preprocessing step to single out outliers or possible sample classes in supervised learning \cite{boongoen2018cluster}.  In real datasets, there is no precise definition of clusters due to the lack of prior knowledge and the clusters may appear with different compactness and separability. Hence, for the same dataset, the clustering algorithms with different objective functions or the same ones with different initializations (parameters) may lead to distinct results. For these different partitions, there is a selection problem: which one among them is an appropriate option? Actually, one can avoid making choice by combining these partitions to generate an integrated result, which has the most agreement with these partitions. We call the set of these partitions \textbf{ensemble} and this method \textbf{clustering ensemble} \cite{strehl2003cluster}. Clustering ensemble is an extension of ensemble learning in unsupervised field. This method has received much attention in recent years because of its robustness compared to a single clustering method, less sensitivity to noise, outliers, or sample fluctuations \cite{nguyen2007consensus, vegapons2011a, wu2018a, boongoen2018cluster, zhang2019weighted}. It has been applied in numerous problem domains, such as recommender system \cite{tsai2012cluster, zheng2013penetrate, logesh2020enhancing}, patient stratification \cite{wang2014breast, liu2017entropy, zhang2019a}, image segmentation \cite{zhang2008spectral} and marketing \cite{kuo2016an}.

The standard clustering ensemble consists of two main steps: generation step and fusion step. In generation step, several partitions are generated by running several different clustering algorithms or the same algorithm with different initialization (parameters), etc., forming initial ensemble. In fusion step, the base partitions are combined by consensus function to obtain the integrated result. All partitions are treated equally. However, studies found that: 1) There are redundancy and noise in the base partitions, which can degrade the ensemble efficiency and the ensemble performance\cite{shi2018transfer}. 2) The base partitions are not mutually independent. Some may be highly correlated or may differ significantly, leading to biased or unstable result\cite{li2008weighted}. To handle these challenges, selective clustering ensemble methods\cite{hadjitodorov2006moderate, fern2008cluster, azimi2009adaptive, hong2009resampling, jia2011bagging, alizadeh2011new, alizadeh2011asymmetric, shi2018transfer, li2018cluster, abbasi2019clustering, naldi2013cluster} and weighted clustering ensemble methods\cite{zhou2006clusterer, li2008weighted, gullo2009diversity, alhichri2014clustering, berikov2017ensemble, yang2016overlapping, yousefnezhad2018woce, unlu2019a, zhang2019weighted} are proposed. \textbf{Selective Clustering Ensemble (SCE)}\cite{fern2008cluster} selects a subset of the initial ensemble before fusion step from two perspectives, diversity and stability. \textbf{Weighted Clustering Ensemble (WCE)} \cite{li2008weighted, zhang2019weighted} does not treat the base partitions equally, but weights them according to diversity and stability. Diversity and stability reflect the degree of dissimilarity and similarity between base partitions, respectively. For SCE and WCE methods, the base partitions with higher diversity and stability are usually more preferable. This is because diversity means information from multiple views and less redundancy, and stability means higher agreement with the other base partitions (this is consistent with the goal of clustering ensemble). However, there is a conflict between diversity and stability. Hence, how to make the trade-off between the two is important to SCE and WCE methods, and the key is how to evaluate the quality of the base partitions and the clusters.

Normalized mutual information (NMI) \cite{strehl2003cluster} is usually employed for evaluation of the quality of base partitions \cite{zhou2006clusterer, fern2008cluster, azimi2009adaptive, jia2011bagging}. However, if a partition is judged as low-quality, all clusters within it will also be judged as low-quality, even though there may be some high-quality ones \cite{abbasi2019clustering}. As a result, these high-quality clusters will be neglected in SCE methods and underestimated in WCE method, which is not reasonable. Hence, the quality of clusters should also be evaluated. Although NMI is one of the most popular indices for evaluation of partitions, it  cannot be used to evaluate the quality of a single cluster of interest. To handle this challenge, efforts have been made to extend NMI, such as Binary-NMI (BNMI) \cite{law2004multiobjective}, MAX \cite{alizadeh2011new}, Alizadeh-Parvin-Moshki-Minaei (APMM) \cite{alizadeh2011asymmetric} and Edited NMI (ENMI) \cite{abbasi2019clustering}, etc. The basic motivation behind the extensions is similar, i.e., transforming the cluster under consideration to a partition form and comparing it with a reference one by applying NMI (or its variants). However, BNMI suffers from symmetric problem, and MAX, APMM and ENMI have context meaning problem \cite{li2018cluster}. In order to solve these problems, Li et al. \cite{li2018cluster} proposed a new cluster evaluation method, which is called Set Matching Degree Evaluation (SME). But it is computationally inefficient. They also proposed a new partition evaluation method, SMEP. But it has the problem of ignoring importance of small clusters, as will be discussed later. To address these problems, we propose a new evaluation method for partitions and clusters using kappa and F-score.

The key to this new method is how to solve the labeling correspondence problem. To address this, Liu et al.\cite{liu2019evaluation} proposed a label alignment method that aligns the labels in the clustering results with the true labels through integer linear programming. After label alignment, one can use kappa and F-score as evaluation metrics. However, this alignment method cannot be applied directly in the clustering ensemble since the true labels are unavailable. Hence, we extend this label alignment method and propose a new SCE method based on the new evaluation method, which is called Diversity-Stability-Kappa-F method (DSKF). It uses kappa to select diverse base partitions and uses F-score to evaluate and weight clusters. The advantages of the proposed methods are revealed by the empirical results.

The rest of this paper is organized as follows: Sect.\ref{related_work} first concisely describes the evaluation problem in clustering ensemble and then reviews some representative evaluation indices in unsupervised fields. Following that, Sect. \ref{sect3} first propose a new evaluation method for partitions and clusters using kappa and F-score, and further propose a new SCE method DSKF. Sect. \ref{sect4} reveals the advantages of the proposed methods. Finally, this paper is concluded in Sect \ref{sect6}.

\begin{figure}[!t]
\centering
\includegraphics[height=4cm,width=8cm]{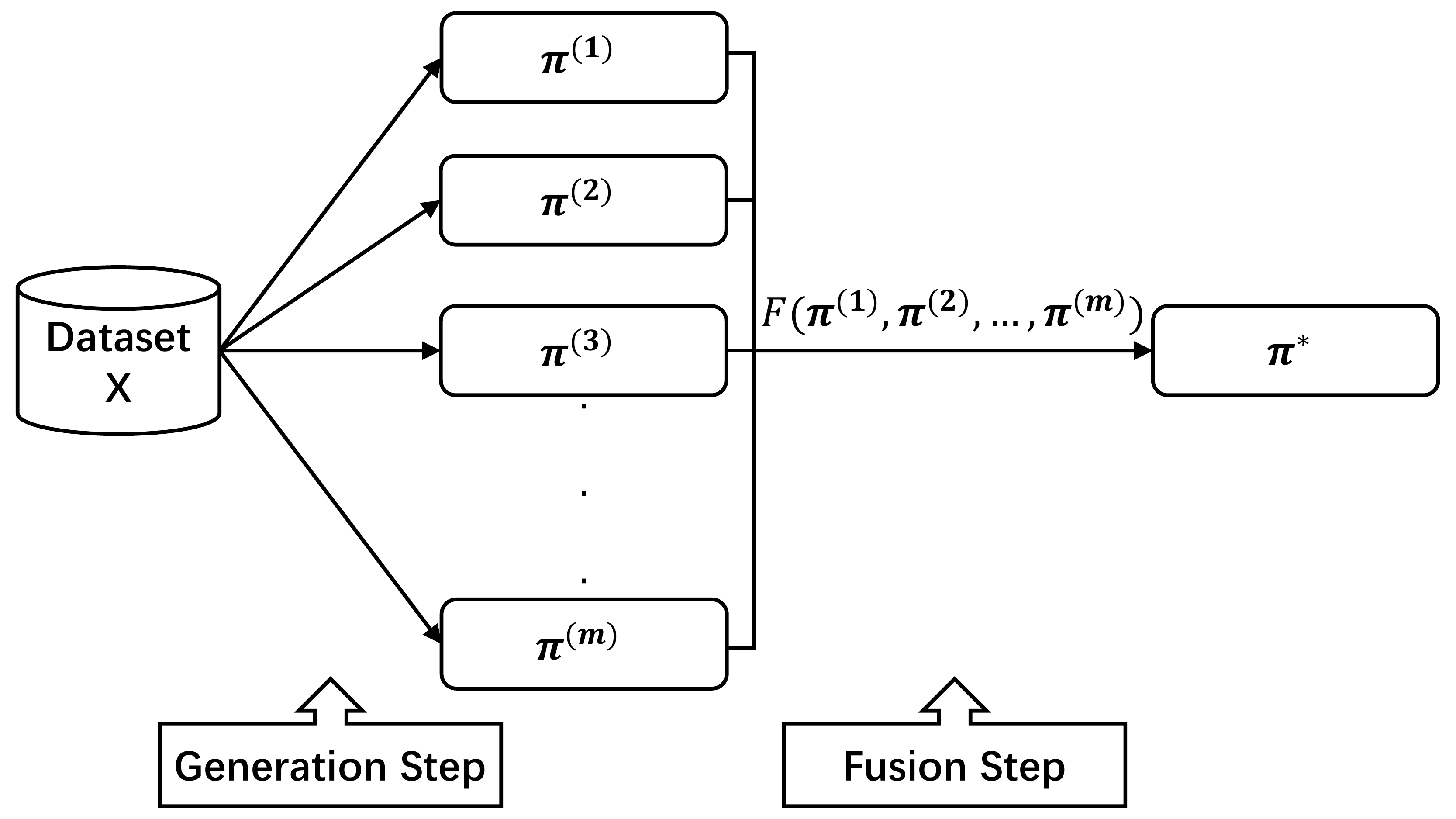}
\caption{The framework of standard clustering ensemble}
\label{fig1}
\end{figure}

\section{Related Work}\label{related_work}

In this section, we will first describe the evaluation problem in clustering ensemble, and then concisely review some representative evaluation indices in unsupervised and supervised fields.

\subsection{The evaluation problem in clustering ensemble}

Clustering ensemble \cite{strehl2003cluster}, also known as consensus clustering \cite{monti2003consensus}, is a method of combining multiple partitions obtained from different clustering methods or from different runs of the same method. It has received much attention in recent years because of its robustness compared to a single clustering algorithm, less sensitivity to noise, outliers, or sample fluctuations \cite{nguyen2007consensus, vegapons2011a, wu2018a, boongoen2018cluster, zhang2019weighted}. The standard clustering ensemble consists of two main steps: generation step and fusion step, as is shown in Fig.\ref{fig1}. In generation step, $m$ base partitions $\pi^{(1)}, \pi^{(2)}, \pi^{(3)}, \cdots, \pi^{(m)}$ are generated by running several different clustering algorithms or the same algorithm with different initializations (parameters), etc., forming initial ensemble. In fusion step, the base partitions are combined by consensus function $F$ to obtain the final result $\pi^{*}$. All partitions are treated equally.

Studies found that: 1) There are redundancy and noise in the base partitions, which can degrade the ensemble efficiency and the ensemble performance \cite{shi2018transfer}. 2) The base partitions are not mutually independent. Some may be highly correlated or may differ significantly, leading to biased or unstable result \cite{li2008weighted}. To handle these challenges, selective clustering ensemble methods \cite{hadjitodorov2006moderate, fern2008cluster, azimi2009adaptive, hong2009resampling, jia2011bagging, alizadeh2011new, alizadeh2011asymmetric, shi2018transfer, li2018cluster, abbasi2019clustering} and weighted clustering ensemble methods \cite{zhou2006clusterer, li2008weighted, berikov2017ensemble, yang2016overlapping, unlu2019a, zhang2019weighted} are proposed. Selective Clustering Ensemble (SCE) \cite{fern2008cluster} selects a subset of the initial ensemble before fusion step from two perspectives, diversity and stability. Weighted Clustering Ensemble (WCE) \cite{li2008weighted, zhang2019weighted} does not treat the base partitions equally, but weights them according to diversity and stability too. In addition to the base partitions, clusters in the partitions are also selected or weighted \cite{alizadeh2011new, alizadeh2011asymmetric, yang2016overlapping, li2018cluster, abbasi2019clustering}. It is obvious that how to make the trade-off between diversity and quality of the base partitions and clusters is important to the SCE and WCE methods. The key point here is how to evaluate the quality of the base partitions and the clusters without prior knowledge. Several criteria have been proposed. A brief summary is given below.

\subsection{Evaluation methods for clustering in unsupervised field}
\label{2.1}
In this subsection, we will begin with normalized mutual information, and briefly review some representative indices for evaluation of partitions and clusters in SCE and WCE.

\subsubsection{Normalized Mutual Information (NMI)}
    NMI\cite{strehl2003cluster} derives from information theory. Given a partition $\pi^{(1)} = \{c_{1}^{(1)}, c_{2}^{(1)},$ $..., c_{k_1}^{(1)}\}$ of $n$ samples consisting of $k_1$ clusters, $c_{s}^{(1)} \left(s = 1, 2, ..., k_1\right)$ is the \textit{s}-th cluster in $\pi^{(1)}$, its entropy is defined as:
\begin{equation}
H(\pi^{(1)}) = -\sum_{s = 1}^{k_1}\frac{n_{s}}{n}\log\frac{n_{s}}{n},
\end{equation}
where $n_{s}=|c_{s}^{(1)}|$. Similarly, the joint entropy of two partitions $\pi^{(1)}$ and $\pi^{(2)}$ is defined as:
\begin{equation}
H(\pi^{(1)}, \pi^{(2)}) = -\sum_{s = 1}^{k_{1}}\sum_{t = 1}^{k_{2}}\frac{n_{st}}{n}\log\frac{n_{st}}{n},
\end{equation}
    and the conditional entropy is defined as:
    \begin{equation}
H(\pi^{(1)}|\pi^{(2)}) = -\sum_{s = 1}^{k_{1}}\sum_{t = 1}^{k_{2}}\frac{n_{st}}{n}\log\frac{n_{st}}{n_{t}},
\end{equation}
    where $n_{st}=|c_{s}^{(1)}\cap c_{t}^{(2)}|$ and $n_{t}=|c_{t}^{(2)}|$.
    The mutual dependence between $\pi^{(1)}$ and $\pi^{(2)}$ can be measured by the mutual information defined as:
\begin{equation}
MI\left(\pi^{(1)}, \pi^{(2)}\right) = H(\pi^{(1)}) - H(\pi^{(1)}|\pi^{(2)}).
\end{equation}
The value of mutual information has no upper bound, making it hard to interpret the result. The normalized version of $MI\left(\pi^{(1)}, \pi^{(2)}\right)$ ranging from 0 to 1 is defined as:
\begin{equation}
NMI\left(\pi^{(1)}, \pi^{(2)}\right) = \frac{MI\left(\pi^{(1)}, \pi^{(2)}\right)}{\sqrt{H(\pi^{(1)})H(\pi^{(2)})}}.
\label{NMI}
\end{equation}

NMI is one of the most popular indices for evaluation of clustering partitions. However, The index cannot be used to evaluate the quality of a single cluster of interest. This is often very important in real applications, such as weighted ensemble clustering. To handle this challenge, four indices extend NMI. The basic motivation behind the extensions is similar, i.e., transforming the cluster under consideration to a partition form and comparing it with a reference one by applying NMI (or its variants).  Note that there are several other problems for NMI, such as ignoring importance of small clusters, finite size effect and violating proportionality assumption \cite{zhang2015evaluating,Lai2016A,liu2019evaluation}. Simple extensions cannot solve them.
\begin{figure}[!t]
\centering
\includegraphics[height=5cm,width=10cm]{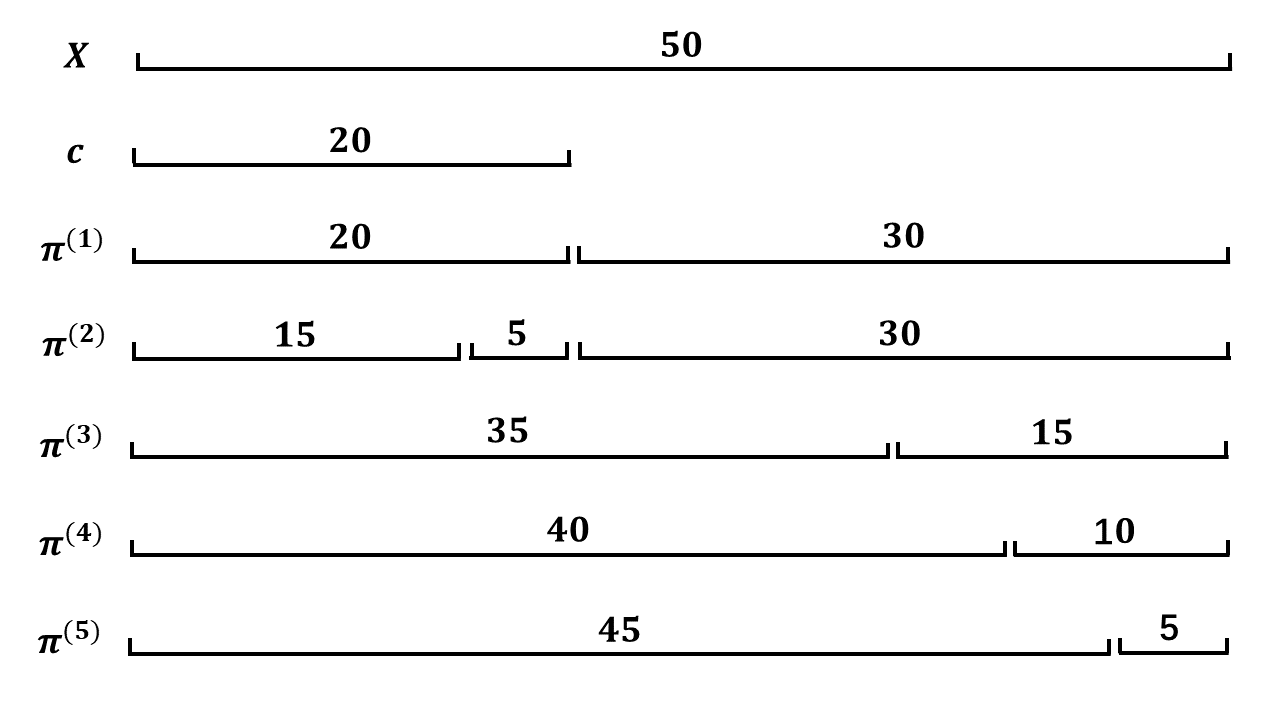}
\caption{An example containing 1 cluster and 5 partitions. The numbers in the figure indicate the number of samples contained in the corresponding set.}
\label{example}
\end{figure}
\begin{table}
\renewcommand{\arraystretch}{1.125} 
\tabcolsep 4mm 
\setlength{\abovecaptionskip}{10pt}
\setlength{\belowcaptionskip}{3pt}%
\caption{The quality of the cluster \textit{c} based on different evaluation methods and different reference partitions. The NMI variants can obtain the same score based on significantly different reference partitions, e.g. $BNMI(c,\,\pi^{(1)}) = BNMI(c,\,\pi^{(2)}), MAX (c,\,\pi^{(4)}) = MAX(c,$ $\pi^{(5)})$, etc. SME and F1 can handle this issue, and their ranks are more intuitive.}
\centering
\begin{tabular}{cccccccc}
\hline\hline
References & BNMI   & MAX & APMM & ENMI & SME    & F1    \\ \hline
$\pi^{(1)}$  & 1    & 1   & 1    & 0.50  & 1      & 1     \\
$\pi^{(2)}$  & 1    & 0.6 & 0.77 & 0.48 & 0.75   & 0.86  \\
$\pi^{(3)}$  & 0.3  & 0.3 & 1    & 0.50  & 0.57   & 0.73  \\
$\pi^{(4)}$  & 0    & 0   & 1    & 0.50  & 0.50    & 0.67  \\
$\pi^{(5)}$  & 0    & 0   & 1    & 0.50  & 0.44   & 0.62  \\
\hline\hline
\label{runtime}
\end{tabular}
\end{table}

\subsubsection{Binary-NMI (BNMI)}
Given a cluster $c$ in a dataset $X$ and the reference partition $\pi$ containing $k$ clusters, BNMI \cite{law2004multiobjective} evaluate the quality of $c$ by converting the cluster-to-partition measure problem into a partition-to-partition measure problem. Firstly, BNMI transforms $c$ to a partition $\pi{'} = \left\{c, X\backslash c\right\}$, where cluster $X\backslash c$ represents the set of samples in dataset \textit{X} that are not in cluster $c$, and changes the reference partition $\pi$ to $\pi{''} = \left\{c^{*}, X\backslash c^{*}\right\}$ accordingly, where $c^{*}$ is the approximation of \textit{c} and is composed of all "positive" clusters in $\pi$. ``Positive'' here means more than half of the samples are in $c$, i.e.,
    \begin{equation}
     c^{*} = \left\{x|x \in c_{t},\,c_{t}\subset\pi,\, |c_{t} \cap c|> \frac{1}{2}|c_{t}|, t = 1, 2, ..., k\right\}.
     \end{equation}
     Then, BNMI is defined as:
\begin{equation}
BNMI\left(c, \pi \right) = NMI\left(\pi{'}, \pi{''} \right).
\label{BNMI}
\end{equation}

We can use the toy example in Fig.\ref{example} to illustrate how to calculate BNMI. There are fifty samples in $X = \{x_i: i=1,\,2,\,\cdots,\,50\}, $ and the cluster $c$ of interest is a subset of $X$, $c = \{x_i: i=1,\,2,\,\cdots,\,20\}$. $\pi^{(1)}$ is the reference partition. Hence, $\pi' = \pi'' = \{\{x_i: i = 1,\,2,\,\cdots,\,20\},\,\{x_i: i = 21,\,22,\,\cdots,\,50\}\}$, meaning that $BNMI(c,\,\pi^{(1)})=1$. Further, one can observe that the quality of $c$ is worse if $\pi^{(2)}$ is the reference partition. However, $BNMI(c,\,\pi^{(2)}) = BNMI(c,\,\pi^{(1)})=1$, which is counterintuitive. This is called symmetric problem \cite{li2018cluster}.

\subsubsection{MAX}

MAX \cite{alizadeh2011new} is proposed to address the problem of BNMI. The only difference between MAX and BNMI is the definition of $c^{*}$. As the name implies, in MAX, $c^{*}$ only consists of the most "positive" cluster rather than all "positive" clusters, i.e., $c^{*} = c_{h}$, where $ h = \arg\max\limits_{t} \left\{|c_{t} \cap c|: c_t\subset\pi,  |c_{t} \cap c|> \frac{1}{2}|c_{t}|, t =\right. \\ \left. 1, 2, ..., k \right\}$. However, although $MAX (c,\,\pi^{(2)}) < MAX(c,\,\pi^{(1)})$, $MAX(c,\pi^{(4)})$ \\ $= MAX(c,$ $\pi^{(5)})=0,$ which is still not reasonable.

\subsubsection{Alizadeh-Parvin-Moshki-Minaei criterion (APMM)}

APMM \cite{alizadeh2011asymmetric} addresses the problem of BNMI in another way. It only considers the samples in $c$, i.e., the samples which are not in $c$ are removed and one can get a new partition on $X\backslash\{x: x\notin c\}$ from $\pi$:
\begin{equation}
\pi{''} = \left\{ \alpha_{t}: \alpha_{t} = c_{t} \cap c, c_t\subset
\pi, c_{t} \cap c \ne \emptyset, t = 1, 2, ..., k\right\}.
\label{cp}
\end{equation}
Then the quality of $c$ is evaluated using $\pi{''}$. However, NMI cannot be applied directly since there is only one cluster in $c$ and $NMI(c, \pi'')$ is always zero no matter what $\pi''$ is.
Hence, APMM modifies NMI and is defined as follows:
\begin{equation}
APMM\left(c, \pi\right) = \frac{2|c|\displaystyle\log{\left(\frac{|c|}{n}\right)}}{|c|\displaystyle\log{\left(\frac{|c|}{n}\right)} + \displaystyle\sum\limits_{i=1}^{k''}|\alpha_{i}|\log{\left(\frac{|\alpha_{i}|}{n}\right)}},
\label{APMM}
\end{equation}
where $k''$ is the number of clusters in $\pi''$. APMM does not solve the problem of MAX, i.e., $APMM(c,\,\pi^{(4)}) = APMM(c,\,\pi^{(5)})$ and $APMM(c,\,\pi^{(1)}) = APMM(c,\,\pi^{(3)})$. This is called context meaning problem \cite{li2018cluster}.

\subsubsection{Edited NMI (ENMI)}

The criterion improves APMM by integrating the samples that are not in $c$ with the partition $\pi''$: $\pi^{\dag}= \pi'' \bigcup \left\{\{x\}| x\in X\backslash c\right\}$. ENMI\cite{abbasi2019clustering} is defined as follows:
\begin{equation}
ENMI\left(c, \pi \right) = NMI\left(\pi{'}, \pi^{\dag} \right),
\label{ENMI}
\end{equation}
where $\pi{'} = \left\{c, X\backslash c\right\}$. ENMI also has the context meaning problem, i.e., $ENMI$ $(c, \pi^{(3)}) = ENMI(c,\,\pi^{(4)})= ENMI(c,\,\pi^{(5)})$.

In order to solve the problems in the NMI-type methods, the method of set matching degree evaluation (SME) was proposed.

\subsubsection{Set Matching Degree Evaluation (SME)}

From the discussions above, one can conclude that NMI and its variants are not suitable for evaluation of the quality of clusters. SME \cite{li2018cluster}
is thus proposed to handle this challenge. SME introduces two concepts, corresponding partition $\pi''$ (already defined in Eq. (\ref{cp})) and extended partition $\pi^{\ddag}$:
\begin{eqnarray}
\pi{''} & = & \left\{ \alpha_{t}: \alpha_{t} = c_{t} \cap c, c_t\subset
\pi, c_{t} \cap c \ne \emptyset, t = 1, 2, ..., k\right\},\\
\pi^{\ddag} & = & \left\{ \beta_{t}: \beta_{t} = c_{t}, c_t\subset
\pi, c_{t} \cap c \ne \emptyset, t = 1, 2, \cdots, k \right\}.
\end{eqnarray}
Obviously, $\pi''$ and $\pi^{\ddag}$ have the same number of clusters according to their definitions, denoted by $k^\ddag$.
The value of SME consists of two parts. The first part evaluates the quality of cluster $c$ using reference partition $\pi''$ and is defined as follows:
\begin{equation}
sim\left(c, \pi''\right) = \frac{1}{|c|}\cdot\max\left\{|\alpha_{i}|,\,i=1,\,2,\,\cdots,\,k^\ddag\right\}.
\label{sme1}
\end{equation}
The second part evaluates the quality of $\pi''$ using partition $\pi^\ddag$ and is defined as follows:
\begin{equation}
sim\left(\pi'', \pi^{\ddag}\right) = \sum\limits_{i = 1}^{k^\ddag}\frac{|\alpha_{i}|}{|c|}\cdot\frac{|\alpha_{i}|}{|\beta_{i}|}.
\label{sme2}
\end{equation}
Combining the two parts, the similarity between $c$ and the reference partition $\pi$ is defined as follows:
\begin{equation}
 SME\left(c, \pi \right) = sim\left(c, \pi''\right)\cdot sim\left(\pi'', \pi^{\ddag}\right).
\label{SME}
\end{equation}

It has been proved that SME has neither symmetric problem nor context meaning problem, and the criterion can be naturally extended to measuring the similarity between two partitions, denoted by SMEP. However, SMEP does not notice the importance of small clusters in partitions. For example, given the reference partition $ [1,\,1,\,1,\,1,\,2]$, the values of SMEP of two computed partitions $[1,\,1,\,1,\,2,\,2]$ and $[1,\,1,\,1,\,1,\,1]$ are 0.55 and 0.65, respectively, which are not reasonable.  In addition, calculating SME \underline{and SMEP} are time consuming, as will be shown in Sect \ref{running_time}.

\subsubsection{Kappa}

The main challenge in evaluation of clustering results is the limited amount of information available\cite{liu2019evaluation}. From the results, one can only know which samples are clustered together and which ones are not, making it difficult to point-wise compare between the ground-truth labels and the computed ones. For example, the two clustering results $[1,\,1,\,1,\,1,\,1,\,1,\,2,\,2,\,3,\,3]$ and $[2,\,2,\,2,\,2,\,2,\,2,\,3,\,3,\,1,\,1]$ are actually identical.
\setlength{\tabcolsep}{4.3mm}{
\begin{table}
\setlength{\abovecaptionskip}{0pt}%
\setlength{\belowcaptionskip}{10pt}%
\caption{Confusion matrix of binary classification}
\centering
\renewcommand{\arraystretch}{1.3} 
\tabcolsep 5.25mm 
\begin{tabular}{ccccc}
\hline\hline
\quad & \quad  & \multicolumn{3}{c}{Predicted class} \\
\hline
\quad & \quad & \quad & 1 & 0\\
\multirow{2}{0.1cm}{\rotatebox{90}{Actual}} & \multirow{2}{0.4cm}{\rotatebox{90}{class}} & 1 & \hspace{6mm}a\hspace{6mm} & b\\
& & 0& \hspace{6mm}c\hspace{6mm} & d\\
\hline\hline
\label{CM}
\end{tabular}
\end{table}}
To handle this issue, Liu et al.\cite{liu2019evaluation} proposed a label alignment method that aligns the labels in the clustering results with the true labels through integer linear programming. After label alignment, one can use kappa and F-score as evaluation metrics, instead of NMI.

Kappa and F-score are defined for evaluation of classification results.
Specifically, for binary classification problem, based on the confusion matrix in Table \ref{CM}, one can obtain the following five indices\cite{powers2011evaluation,metz1978basic,galton1892}:
\begin{eqnarray}
Accuracy  & = & \frac{a+d}{a + b + c + d}, \\[3mm]
Kappa     & = &\frac{Accuracy - \mathbb{E}\left(Accuracy\right)}{1 - \mathbb{E}\left(Accuracy\right)},\\
Recall    & = & \frac{a}{a + b}, \label{recall} \\[3mm]
Precision & = & \frac{a}{a + c}, \label{precision} \\[3mm]
F\mbox{-}score & = & (1 + \beta^2)\cdot\frac{Precision\cdot Recall}{\left(\beta^2\cdot Precision\right) + Recall},
\end{eqnarray}
where $\displaystyle\mathbb{E}\left(Accuracy\right) = \frac{a + c}{a + b + c + d}\cdot \frac{a + b}{a + b + c + d} + \frac{b + d}{a + b + c + d}\cdot \frac{c + d}{a + b + c + d}$.
\vspace{3mm}
The default value of $\beta$ is 1, and F-score is denoted as F.

\begin{figure}[ht]
\centering
\includegraphics[height=5cm,width=10cm]{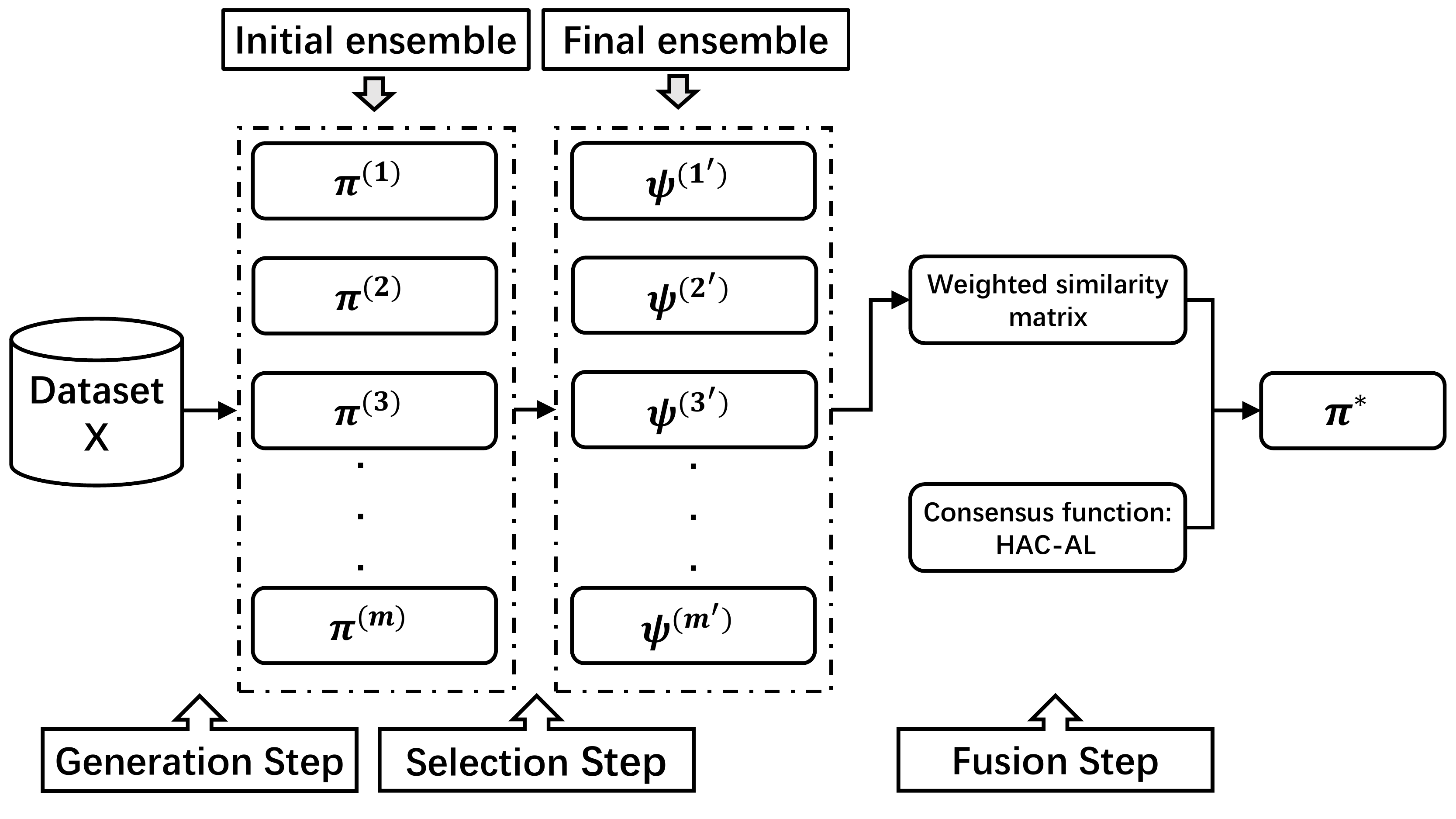}
\caption{The framework of DSKF proposed. The Hierarchical agglomerative clustering with average linkage (HAC-AL) is used as the consensus function to generate the final partition $\pi^{*}$. \textit{m} is the number of partitions in the initial ensemble and $m'$ is the number of partitions in the final ensemble. }
\label{fig2}
\end{figure}

\section{Diversity-Stability-Kappa-F1 method (DSKF)}
\label{sect3}

In this section, we first extend the label alignment method in \cite{liu2019evaluation} to the ensemble field. Then, we propose a new cluster evaluation model and a new SCE framework.

\subsection{Label alignment method extended}
\label{3.1}

Based on the discussion in Sect. \ref{related_work}, we know that the NMI-type methods have several problems, such as symmetric problem, context meaning problem, etc. SME and SMEP are computationally inefficient since they may go through every cluster in the reference partition for evaluation of a single cluster and the number of clusters that need to be evaluated is usually very large. In this section, we propose a new evaluation method for partitions and clusters using kappa and F-score.

We first select a reference partition, and then align all of the base partitions to it through integer linear programming. The key to this alignment method is how to select the appropriate reference partition, since the ground-truth one is not available. In this paper, we just select the reference partition randomly. We name this selection strategy as \textbf{random selection strategy}. Empirical results indicate that the strategy of random selection is effective and efficient.
Specifically, given the reference partition $\pi = \left\{c_{1}, c_{2}, ..., c_{k}\right\}$ and a base partition $\pi^{(i)} = \left\{c_{1}^{(i)}, c_{2}^{(i)}, ..., c_{k_i}^{(i)}\right\}$  under the assumption that $k_i \leq k$, we align the labels in $\pi^{(i)}$ with the ones in $\pi$. The decision variables $y_{st}$ and cost of the alignment $l_{st}$ are defined as follows:
\begin{eqnarray}
y_{st}
& = &
\begin{cases}
1, & \mbox{label $s$ in $\pi^{(i)}$ is aligned to label $t$ in $\pi$}\\[1mm]
0, & \mbox{otherwise}
\end{cases},\\[3mm]
l_{st} & = & |c_{s}^{(i)}\cup c_{t}|-|c_{s}^{(i)}\cap c_{t}|.
\end{eqnarray}
Then, the integer linear programming model can be described as\cite{liu2019evaluation}:
\begin{equation}
\begin{array}{rl}
\min & z = \displaystyle\sum\limits_{s=1}^{k_{i}}\sum\limits_{t=1}^{k}(l_{st}\cdot y_{st})\\[7mm]
s.t. & \left\{
           \begin{array}{l}
             \displaystyle\sum\limits_{s=1}^{k_i}y_{st}\leq1, \, t=1,\, 2,\, \cdots,\, k.\\[7mm]
             \displaystyle\sum\limits_{t=1}^{k}y_{st} = 1, \, s=1,\, 2,\, \cdots,\, k_{i}.\\[7mm]
             y_{st} = 0\ or\ 1.
           \end{array}
    \right.
\end{array}
\label{label alignment}
\end{equation}
Liu et al. \cite{liu2019evaluation} has proved that this model can be transformed into a standard assignment problem and can be solved by Hungarian method. Moreover, it is easy to extend this model to the case $k_i > k$ by replacing the first two constraints in Eq.\ref{label alignment} with $\sum\limits_{s=1}^{k_i}y_{st} = 1$ and $\sum\limits_{t=1}^{k}y_{st} \leq 1$, respectively.

\subsection{Diversity-Stability-Kappa-F1 method (DSKF)}
\label{3.2}
In selective clustering ensemble methods (SCE) and weighted clustering ensemble methods (WCE), the base partitions and the clusters with higher diversity and stability are usually more preferable. This is because diversity means information from multiple views and less redundancy, and stability means higher agreement with the other base partitions (this is consistent with the goal of clustering ensemble). However, there is a conflict between diversity and stability, and how to balance between the two is a challenge problem.

To handle this challenge, one can utilize diversity and stability in turn in different stages \cite{li2018cluster}. For the selection step, its main function is to select base partitions containing information from multiple views, hence, diversity should be dominant in this step. For the fusion step, the goal is to seek result that has the most agreement, hence, stability should play a major role in this step. We propose Diversity-Stability-Kappa-F method (DSKF), which is shown in Fig.\ref{fig2}. It uses kappa to select diverse base partitions and uses F-score to give more weight to the more stable clusters. Details are given below.

\subsubsection{Generation and selection step}

\label{selection stragy}
Firstly, we use K-means (KM) algorithm \cite{macqueen1967some} with different initializations to generate $m$ base partitions forming the initial ensemble $\Pi$, where $\Pi = \left\{\pi^{(1)}, \pi^{(2)}, \cdots, \pi^{(m)}\right\}$. Then we align the labels among the partitions through integer linear programming, forming the aligned initial ensemble $\Psi = \{\psi^{(1)},$ $\psi^{(2)},\cdots, \psi^{(m)}\}$, where $\psi^{(i)}$ is the aligned partition that corresponds to $\pi^{(i)} (i = 1, 2, \cdots, m)$.  After label alignment, one can use kappa to measure the diversity of the partitions:
\begin{equation}
D\left(\psi^{(i)}\right) = 1-\frac{1}{m-1}\sum\limits_{j \ne i}^{m}kappa\left(\psi^{(i)}, \psi^{(j)}\right),\, i =1, 2, \cdots, m.
\label{Di}
\end{equation}
The aligned partitions will be selected if their diversity are more than a threshold $\sigma$ we given. Finally, we obtain $m'$ base partitions having high diversity, denoted by $\Psi^{'}$.  The partition-based denotation of $\Psi^{'}$ is $\left\{\psi^{(1^{'})}, \psi^{(2^{'})}, \cdots,\right. \\ \left. \psi^{(m{'})}\right\} $, and the cluster-based denotation is $\left\{c_{1}, c_{2}, \cdots, c_{h}\right\}$, where $m{'}$ is the number of partitions in $\Psi^{'}$ and \textit{h} is the number of clusters included in all partitions in $\Psi^{'}$.

\subsubsection{Fusion step}

In order to combine the aligned partitions in final ensemble $\Psi^{'}$ by treating the clusters in these partitions unequally, the binary cluster-association matrix $B^{(n\times h)}$ \cite{iamon2011a} and the weighted similarity matrix (or, weighted co-association matrix) $S^{(n\times n)}$ \cite{zhang2019weighted} are introduced in this step, where $n$ is the number of samples and $h$ is the number of clusters included in all partitions in $\Psi^{'}$.

$B$ indicates the membership of samples and is defined as follows:
$$B(i,\, j) =\begin{cases} 1,  & x_i \in c_j\\ 0, & x_i \notin c_j, \end{cases}$$ indicating whether the $i$th sample $x_i$ is in the $j$th cluster $c_j$.

$S$ reveals the weighted frequency that two samples are assigned into the same cluster. To construct the matrix $S$, we first evaluate the quality of clusters with F. Given a cluster $c_{i}$, $c_{i}\in \psi^{(p{'})} \left(1 \leq p{'} \leq m{'}\right)$, its quality can be calculated as:
\begin{equation}
l_{i}= \frac{1}{m{'}-1}\sum\limits_{j{'} \ne p{'}}^{m{'}}F\left(c_{i}, \psi^{(j{'})}\right), i = 1, 2, \cdots, h.
\end{equation}
$l$ is normalized to improve its interpretability: $w_{i} = \dfrac{l_{i}}{\sum_{j = 1}^{h}l_{j}}, i = 1, 2, \cdots, h$, making up a diagonal matrix $W^{(h\times h)} = diag\left(w_{1}, w_{2}, \cdots, w_{h}\right)$, counting for the quality of clusters.  Then, $S$ can be defined as follows:
\begin{equation}
S = BWB^{T}.
\label{WCO}
\end{equation}

Finally, one can use hierarchical agglomerative clustering method with average linkage (HAC-AL) \cite{johnson1967hierarchical} on this matrix to generate the final result $\pi^{*}$.

\begin{table*}[!t]
\setlength{\abovecaptionskip}{0pt}%
\setlength{\belowcaptionskip}{10pt}%
\caption{Description of datasets.}
\resizebox{\textwidth}{46mm}{
\renewcommand{\arraystretch}{1.25} 
\tabcolsep 0.8mm 
\begin{tabular}{ccccc}
\hline\hline
\textbf{Dataset label} & \textbf{Dataset name} & \textbf{No. of samples (n)} & \textbf{No. of features} & \textbf{No. of classes $\left(k^{*}\right)$}\\\hline
\textbf{1} & Iris                                & 150    & 4   & 3 \\
\textbf{2} & Wine                                & 178    & 13  & 3 \\
\textbf{3} & Seeds                               & 210    & 7   & 3 \\
\textbf{4} & Glass                               & 214    & 9   & 6 \\
\textbf{5} & Protein Localization Sites          & 272    & 7   & 3 \\
\textbf{6} & Ecoli                               & 336    & 7   & 8 \\
\textbf{7} & LIBRAS Movement Database            & 360    & 90  & 15 \\
\textbf{8} & User Knowledge Modeling             & 403    & 5   & 4 \\
\textbf{9} & Vote                                & 435    & 16  & 2 \\
\textbf{10} & Wisconsin Diagnostic Breast Cancer & 569    & 30  & 2 \\
\textbf{11} & Synthetic Control Chart Time Series& 600    & 60  & 6 \\
\textbf{12} & Australian Credit Approval         & 690    & 14  & 2 \\
\textbf{13} & Cardiotocography                   & 2126   & 40  & 10 \\
\textbf{14} & Wave form Database Generator       & 5000   & 21  & 3 \\
\textbf{15} & Parkinsons Telemonitoring          & 5875   & 21  & 42 \\
\textbf{16} & Statlog Landsat Satellite          & 6435   & 36  & 6 \\
\textbf{17} & Tr12                               & 313    & 5804   & 8 \\
\textbf{18} & Tr11                               & 414    & 6428   & 9 \\
\textbf{19} & Tr45                               & 690    & 8261   & 10 \\
\textbf{20} & Tr41                               & 878    & 7454   & 10 \\
\textbf{21} & Tr31                               & 927    & 10128  & 7 \\
\textbf{22} & Wap                                & 1560   & 8460   & 20 \\
\textbf{23} & Hitech                             & 2301   & 126321 & 6 \\
\textbf{24} & Fbis                               & 2463   & 2000   & 17 \\
\hline\hline
\label{datasets}
\end{tabular}}
\end{table*}

\section{Experimental results}
\label{sect4}
In this section, we validate the effectiveness and the efficiency of our method on multiple real world datasets.

\subsection{Experimental Settings}

We select 16 real datasets from the UCI Machine Learning Repository (UCI, http://archive.ics.uci.edu/ml/) and 8 document datasets from the C-LUTO clustering toolkit \cite{steinbach2000a} for our experiments. The basic information is listed in Table \ref{datasets}.

The parameter settings are the same as in the literature \cite{li2018cluster}. Specifically: 1) In the generation step, K-means is used to generate 50 base partitions. Euclidean distance is employed for the UCI real datasets and cosine similarity is employed for the document datasets\cite{li2018cluster}. 2) The number of clusters is set to a random number in the range of $[k^{*},\, \lfloor\sqrt{n}\rfloor]$ for the UCI real datasets \cite{kuncheva2004using}, where $k^{*}$ is the true number of clusters. The number of clusters is equal to $k^{*}$ for the document datasets \cite{xu2016an}. 3) In the selection step, 25 aligned partitions are selected to form the final ensemble. 4) In the fusion step, the number of clusters is set to $k^{*}$. To eliminate the randomness effect of the initial ensemble, we report the average scores of NMI and kappa over 50 independent runs. In addition, all scores in tables are shown to two decimal places only due to space limitations.

\begin{figure}
\centering
\includegraphics[height=70mm,width=140mm]{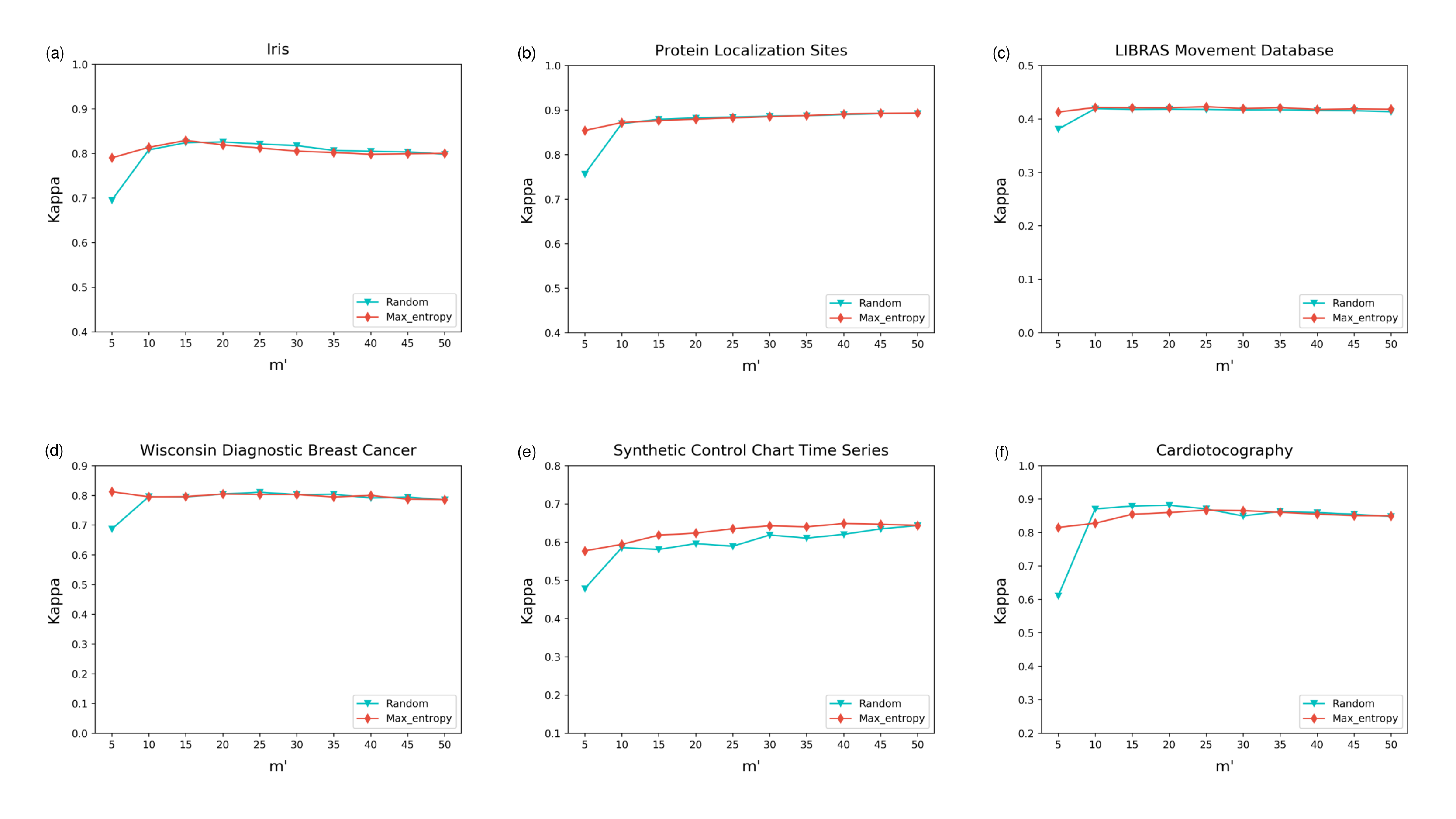}
\caption{The performance of different selection strategies for the reference partition. The size of the final ensemble $m^{'}$ ranges from 5 to 50 with an interval of 5. "Random" denotes the performance of DSKF with the random selection strategy. "Max\_entropy" denotes the performance of DSKF with the maximum entropy principle. The values are averages of 50 trials.}
\label{fig4}
\end{figure}

\begin{figure}
\centering
\includegraphics[height=70mm,width=140mm]{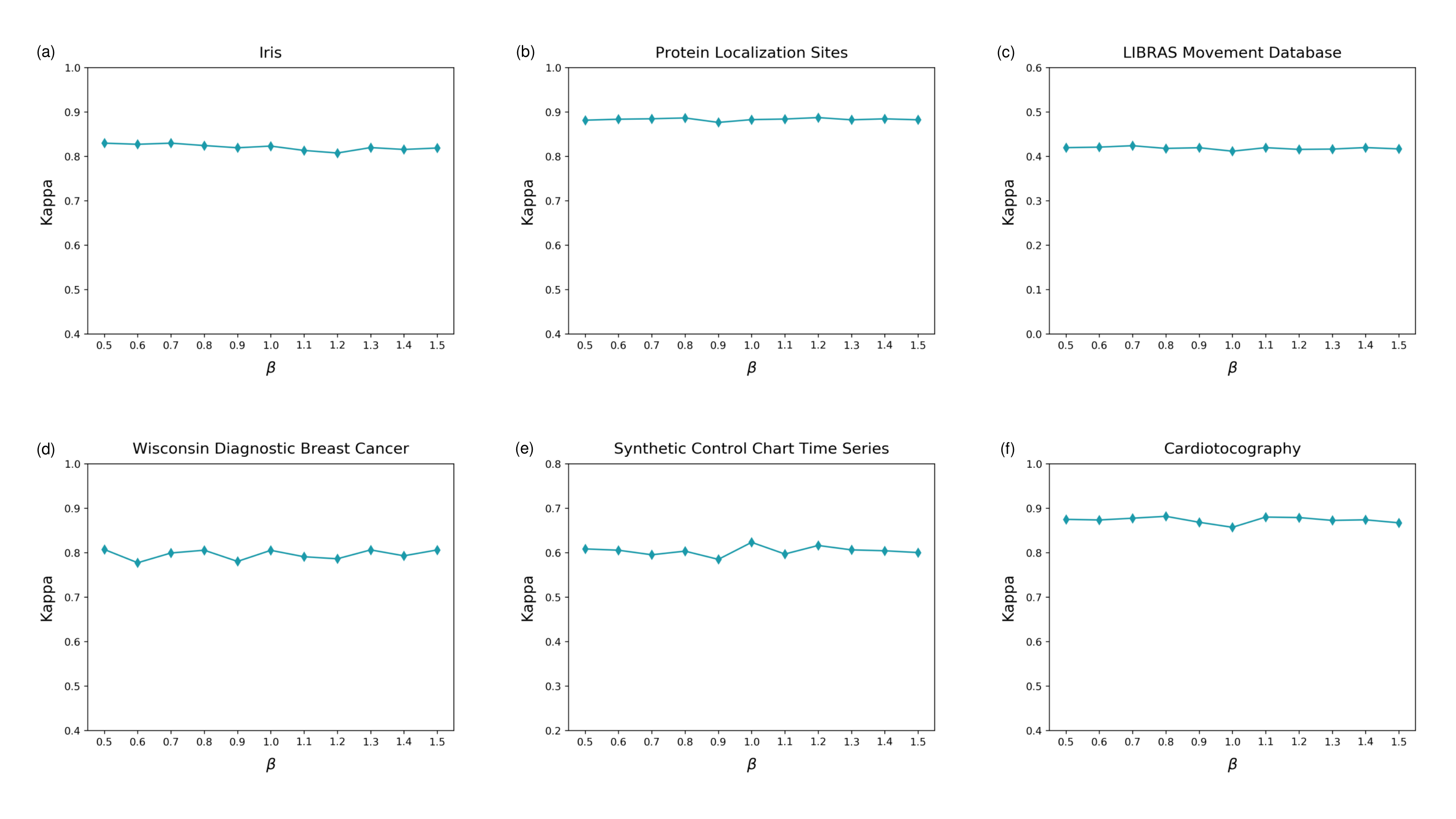}
\caption{The relations between $\beta \in \left\{0.5,\, 0.6,\, \cdots,\, 1.5\right\}$ and the performance of DSKF. The values are averages of 50 trials.}
\label{fig5}
\end{figure}

\subsection{Effectiveness analysis of the random selection strategy and sensitivity analysis of ensemble performance to $\beta$}
\label{5.2}
In the proposed method, the reference partition for label alignment is selected randomly, whose quality cannot be guaranteed. A low-quality reference partition may be obtained, and the ensemble performance may be degraded by this. In addition, the value of $\beta$ in the F-score also has an impact on ensemble performance. Therefore, we conducted experiments over six real datasets to validate the effectiveness of the random selection strategy and the sensitivity of ensemble performance to $\beta$.

Firstly, we compare the random selection strategy with maximum entropy selection strategy \cite{ayad2010on, alhichri2014clustering}. In this strategy, the base partition with the maximum entropy in the initial ensemble is used as the reference partition. The results are shown in Fig.\ref{fig4}, from which one can observe that: 1) The two strategies are of about the same strength, meaning that the random selection strategy is enough. 2). The methods do not achieve their best results over full ensemble and are benefited from the selection and the weighting steps.

In addition, Fig.\ref{fig5} analyzes the relations between $\beta$ in F-score and the performance of DSKF. From the figure, one can observe that the ensemble performance is insensitivity to the parameter $\beta$ and $\beta = 1$ is enough.

\subsection{Performance analysis of the proposed evaluation method and the proposed DSKF}
\label{5.2}

In this subsection, we evaluate the effectiveness and the efficiency of the proposed evaluation method and the DSFK method.

\subsubsection{The effectiveness analysis}
Firstly,  to verify the performance of kappa in selecting base partitions and F in weighting clusters, we use several state-of-the-art evaluation methods as baselines, including NMI, ENMI, SME and SMEP. Specifically, we use NMI, SMEP or kappa to select diverse base partitions and use four standard clustering ensemble methods to integrate these selected partitions, which are WCT+KM \cite{iamon2011a}, WTQ+KM \cite{iamon2011a}, CSPA \cite{strehl2003cluster} and EAC-AL \cite{fred2005combining}. Similarly, we use ENMI, SME or F to weight clusters in base partitions and integrate the weighted base partitions using the above four ensemble methods.

Secondly, to verify the effectiveness of the proposed DSKF method, we use
four state-of-the-art clustering ensemble methods as baselines, including three standard clustering ensemble methods (CESHL \cite{zhou2022clustering}, SPCE \cite{zhou2020self} and TRCE \cite{zhou2021tri}) and one selective clustering ensemble method Diversity-Stability-SMEP-SME (DSME) \cite{li2018cluster}. The DSME  uses SMEP to select diverse base partitions and uses SME to weight clusters. The ensemble performance is shown in Tables \ref{select_nmi}-\ref{ce_kappa}.

From the tables, one can observe that: 1) Based on NMI, the results of ensemble methods do not show a significant improvement and sometimes even regress. The ranks of the methods based on kappa are different from those based on NMI. All of these differences are due to the drawbacks of NMI, such as the finite size effect, preferring large number of clusters. For example, on the dataset \textit{Tr12} ($n = 313, k^* = 8$), given a set of base partitions, the number of clusters in final partition automatically chosen by SPCE with $\gamma = 0.2$ is 8 (true cluster number) and the NMI is 0.65. The $\gamma$ is a hyperparameter in SPCE.  However, when $\gamma = 0.9$, the number of clusters is 164, and NMI is still 0.65! On the contrary, the kappa value of $k = 8$ is 0.56, and that of $k = 164$ is 0.30. This rank is more reasonable. Similarly, on the dataset {\it Parkinsons Telemonitoring}, the NMI of K-means with $k = 42$ (true cluster number) is 0.67, and that of K-means with $k = 80$ is 0.72. The kappa value of $k = 42$ is 0.37, and that of $k = 80$ is 0.34. 2) The results of the proposed methods are among the best based on kappa.

\begin{figure*}
\centering
\includegraphics[height=50mm,width=100mm]{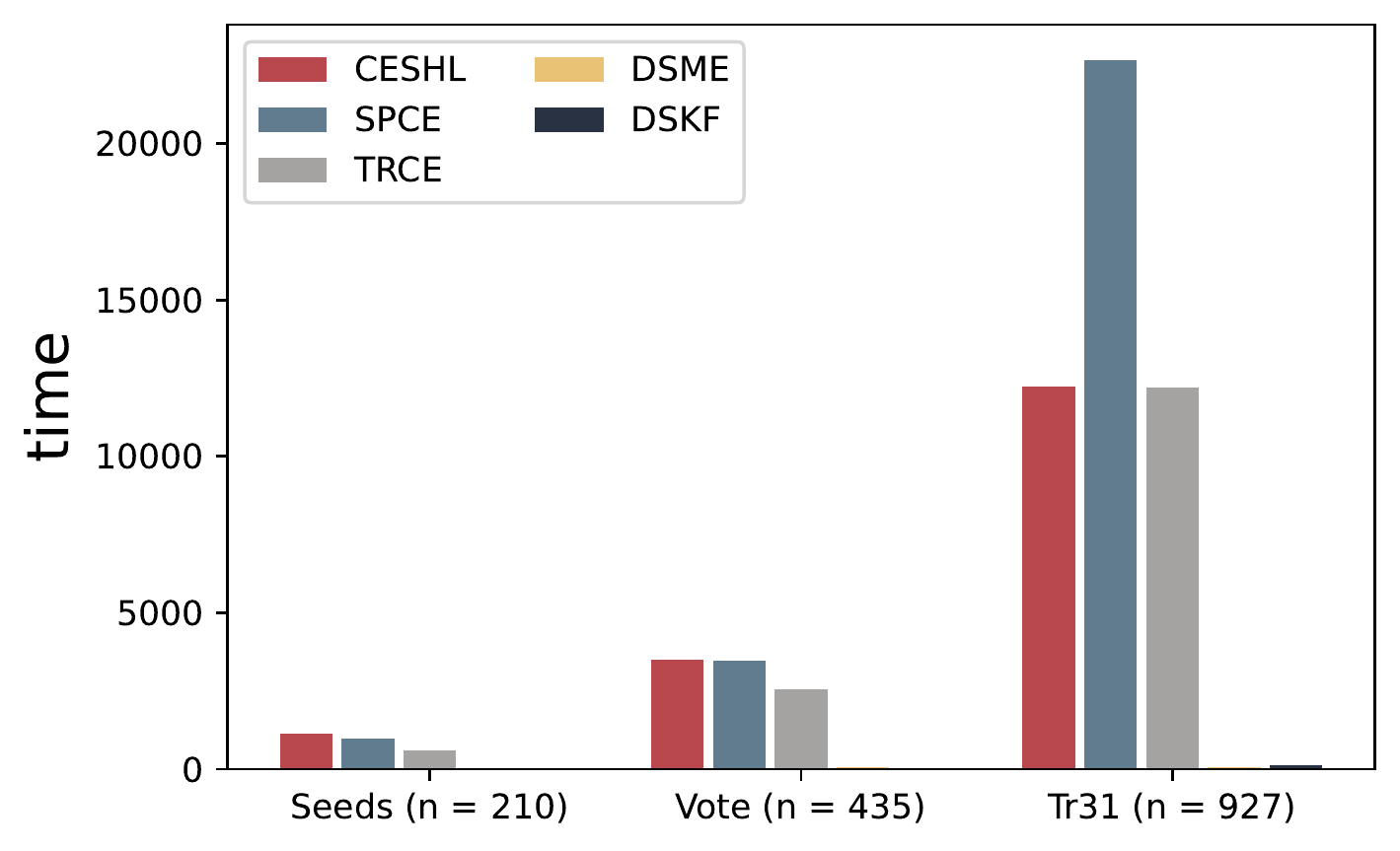}
\caption{Comparison of running time (in seconds) of different methods on three datasets}
\label{run_time}
\end{figure*}

\begin{table*}
\setlength{\abovecaptionskip}{0pt}%
\setlength{\belowcaptionskip}{10pt}%
\caption{NMI of the selective clustering ensemble methods with different evaluation methods. For each comparison, the best result is highlighted in boldface. The last line displays the sum of the number of highlighted values in each column, which is denoted as COUNT. Note that the statistics of base partitions are presented in the second and third columns.}
\scalebox{0.55}{
\renewcommand{\arraystretch}{0.9} 
\tabcolsep 2.25mm 
\begin{tabular}{ccccccccccccccc}
\hline\hline
\multirow{2}{*}{Dataset}&\multicolumn{2}{c}{KM}
                        &\multicolumn{3}{c}{WCT+KM} &\multicolumn{3}{c}{WTQ+KM}
                        &\multicolumn{3}{c}{CSPA}&\multicolumn{3}{c}{EAC-AL}\\
                        \cmidrule(r){2-3}
                        \cmidrule(r){4-6} \cmidrule(r){7-9}
                        \cmidrule(r){10-12} \cmidrule(r){13-15}
\quad                   &Avg(base) &Std(base)
                        &NMI &SMEP &Kappa        &NMI &SMEP &Kappa
                        &NMI &SMEP &Kappa        &NMI &SMEP &Kappa\\\hline
1  &0.65 &0.04 &0.77 &\textbf{0.78} &0.74 &\textbf{0.75} &0.74 &0.74 &0.81 &0.81 &\textbf{0.84} &0.78 &\textbf{0.79} &0.76 \\
2  &0.69 &0.07 &\textbf{0.86} &0.85 &0.85 &\textbf{0.85} &0.84 &0.85 &0.79 &\textbf{0.80} &0.79 &\textbf{0.87} &0.85 &0.86 \\
3  &0.57 &0.04 &0.67 &\textbf{0.68} &0.67 &0.68 &\textbf{0.68} &0.68 &0.73 &0.74 &\textbf{0.74} &0.71 &\textbf{0.72} &0.70 \\
4  &0.36 &0.03 &0.34 &\textbf{0.34} &0.34 &0.33 &\textbf{0.33} &0.33 &\textbf{0.31} &0.30 &0.28 &\textbf{0.34} &0.34 &0.33 \\
5  &0.57 &0.06 &\textbf{0.71} &0.70 &0.71 &\textbf{0.73} &0.71 &0.72 &0.46 &\textbf{0.47} &0.46 &0.74 &\textbf{0.75} &0.74 \\
6  &0.58 &0.02 &0.59 &0.59 &\textbf{0.59} &0.58 &0.58 &\textbf{0.59} &\textbf{0.52} &0.52 &0.52 &0.63 &0.62 &\textbf{0.63} \\
7  &0.62 &0.02 &0.6 &\textbf{0.60} &0.60 &0.60 &\textbf{0.61} &0.60 &0.56 &\textbf{0.57} &0.57 &0.61 &\textbf{0.62} &0.61 \\
8  &0.33 &0.04 &0.37 &\textbf{0.39} &0.36 &0.38 &\textbf{0.38} &0.37 &0.36 &0.37 &\textbf{0.38} &0.36 &\textbf{0.40} &0.35 \\
9  &0.40 &0.04 &\textbf{0.49} &0.49 &0.48 &0.47 &\textbf{0.47} &0.47 &0.43 &\textbf{0.44} &0.44 &0.48 &0.43 &\textbf{0.49} \\
10 &0.42 &0.04 &0.60 &\textbf{0.64} &0.61 &0.61 &\textbf{0.64} &0.61 &0.45 &0.47 &\textbf{0.48} &0.54 &\textbf{0.58} &0.51 \\
11 &0.73 &0.03 &0.78 &\textbf{0.78} &0.77 &0.79 &\textbf{0.79} &0.79 &0.80 &\textbf{0.80} &0.79 &0.78 &\textbf{0.78} &0.78 \\
12 &0.24 &0.02 &\textbf{0.28} &0.27 &0.22 &0.20 &\textbf{0.22} &0.17 &\textbf{0.33} &0.32 &0.31 &0.34 &\textbf{0.35} &0.32 \\
13 &0.82 &0.05 &0.86 &0.85 &\textbf{0.87} &0.87 &0.85 &\textbf{0.87} &0.76 &\textbf{0.76} &0.76 &\textbf{0.96} &0.95 &0.95 \\
14 &0.37 &0.04 &0.37 &0.37 &\textbf{0.38} &0.37 &0.37 &\textbf{0.37} &0.35 &0.37 &\textbf{0.39} &0.39 &0.41 &\textbf{0.42} \\
15 &0.70 &0.01 &0.69 &\textbf{0.70} &0.69 &0.69 &\textbf{0.69} &0.69 &0.67 &\textbf{0.67} &0.67 &0.70 &\textbf{0.70} &0.70 \\
16 &0.55 &0.03 &\textbf{0.61} &0.60 &0.60 &0.58 &\textbf{0.59} &0.58 &\textbf{0.54} &0.53 &0.54 &0.64 &0.64 &\textbf{0.64} \\
17 &0.56 &0.07 &0.65 &0.65 &\textbf{0.66} &0.65 &0.65 &\textbf{0.66} &0.59 &0.60 &\textbf{0.61} &0.65 &0.65 &\textbf{0.67} \\
18 &0.62 &0.05 &0.71 &0.71 &\textbf{0.73} &0.69 &0.69 &\textbf{0.71} &0.61 &0.61 &\textbf{0.62} &0.72 &0.73 &\textbf{0.74} \\
19 &0.63 &0.05 &0.65 &0.66 &\textbf{0.68} &0.65 &0.66 &\textbf{0.67} &0.59 &0.60 &\textbf{0.62} &0.68 &0.70 &\textbf{0.71} \\
20 &0.61 &0.04 &0.67 &\textbf{0.67} &0.67 &0.66 &\textbf{0.67} &0.67 &0.61 &0.62 &\textbf{0.62} &0.67 &0.67 &\textbf{0.68} \\
21 &0.52 &0.04 &0.54 &\textbf{0.54} &0.52 &0.53 &\textbf{0.54} &0.52 &\textbf{0.50} &0.49 &0.48 &0.56 &\textbf{0.58} &0.55 \\
22 &0.54 &0.02 &0.61 &0.61 &\textbf{0.62} &0.60 &0.60 &\textbf{0.60} &0.57 &0.57 &\textbf{0.57} &0.63 &0.63 &\textbf{0.63} \\
23 &0.31 &0.02 &\textbf{0.34} &0.34 &0.34 &0.33 &\textbf{0.33} &0.32 &\textbf{0.33} &0.33 &0.32 &0.34 &\textbf{0.34} &0.34 \\
24 &0.58 &0.02 &0.59 &0.59 &\textbf{0.60} &0.58 &0.58 &\textbf{0.59} &0.55 &0.55 &\textbf{0.55} &0.61 &\textbf{0.61} &0.61 \\
\hline
COUNT &-&- &6&10&8 &3&13&8 &6&7&11 &3&12&9\\
\hline\hline
\end{tabular}\label{select_nmi}}
\end{table*}

\begin{table*}
\setlength{\abovecaptionskip}{1pt}%
\setlength{\belowcaptionskip}{10pt}%
\caption{Kappa of the selective clustering ensemble methods with different evaluation methods. For each comparison, the best result is highlighted in boldface. The last line displays the sum of the number of highlighted values in each column, which is denoted as COUNT. Note that the statistics of base partitions are presented in the second and third columns.}
\scalebox{0.55}{
\renewcommand{\arraystretch}{0.9} 
\tabcolsep 2.25mm 
\begin{tabular}{ccccccccccccccc}
\hline\hline
\multirow{2}{*}{Dataset}&\multicolumn{2}{c}{KM}
                        &\multicolumn{3}{c}{WCT+KM} &\multicolumn{3}{c}{WTQ+KM}
                        &\multicolumn{3}{c}{CSPA}&\multicolumn{3}{c}{EAC-AL}\\
                        \cmidrule(r){2-3}
                        \cmidrule(r){4-6} \cmidrule(r){7-9}
                        \cmidrule(r){10-12} \cmidrule(r){13-15}
\quad                   &Avg(base) &Std(base)
                        &NMI &SMEP &Kappa        &NMI &SMEP &Kappa
                        &NMI &SMEP &Kappa        &NMI &SMEP &Kappa\\\hline
1  &0.42 &0.12 &0.77 &\textbf{0.78} &0.75 &0.72 &0.71 &\textbf{0.73} &0.90 &0.90 &\textbf{0.93} &0.82 &\textbf{0.83} &0.80 \\
2  &0.54 &0.17 &0.91 &0.90 &\textbf{0.91} &\textbf{0.91} &0.90 &0.91 &0.89 &\textbf{0.89} &0.89 &\textbf{0.95} &0.94 &0.94 \\
3  &0.36 &0.14 &\textbf{0.80} &0.79 &0.76 &0.79 &\textbf{0.80} &0.78 &0.88 &0.88 &\textbf{0.89} &0.85 &\textbf{0.85} &0.84 \\
4  &0.20 &0.05 &0.27 &\textbf{0.27} &0.27 &0.27 &\textbf{0.27} &0.27 &\textbf{0.32} &0.32 &0.29 &\textbf{0.30} &0.30 &0.27 \\
5  &0.39 &0.20 &\textbf{0.82} &0.79 &0.81 &\textbf{0.84} &0.80 &0.81 &\textbf{0.57} &0.57 &0.56 &0.89 &0.89 &\textbf{0.89} \\
6  &0.38 &0.06 &0.46 &0.45 &\textbf{0.46} &0.44 &0.44 &\textbf{0.46} &\textbf{0.42} &0.42 &0.42 &0.53 &0.51 &\textbf{0.54} \\
7  &0.33 &0.04 &0.41 &\textbf{0.41} &0.40 &0.40 &\textbf{0.40} &0.40 &0.42 &\textbf{0.43} &0.42 &0.41 &\textbf{0.41} &0.41 \\
8  &0.20 &0.07 &0.37 &\textbf{0.38} &0.37 &0.37 &\textbf{0.37} &0.37 &0.40 &\textbf{0.41} &0.40 &0.38 &\textbf{0.40} &0.38 \\
9  &0.29 &0.13 &\textbf{0.76} &0.75 &0.74 &0.73 &\textbf{0.73} &0.72 &0.69 &\textbf{0.70} &0.70 &\textbf{0.74} &0.70 &0.74 \\
10 &0.21 &0.14 &0.84 &\textbf{0.86} &0.84 &0.84 &\textbf{0.86} &0.84 &0.69 &0.71 &\textbf{0.71} &0.79 &\textbf{0.81} &0.76 \\
11 &0.49 &0.12 &0.64 &0.64 &\textbf{0.64} &0.63 &0.64 &\textbf{0.64} &0.82 &\textbf{0.82} &0.81 &0.59 &\textbf{0.64} &0.61 \\
12 &0.16 &0.09 &\textbf{0.53} &0.52 &0.45 &0.42 &\textbf{0.45} &0.38 &\textbf{0.62} &0.61 &0.61 &0.63 &\textbf{0.64} &0.58 \\
13 &0.49 &0.14 &0.68 &0.67 &\textbf{0.69} &0.68 &0.66 &\textbf{0.69} &\textbf{0.55} &0.55 &0.55 &\textbf{0.90} &0.87 &0.87 \\
14 &0.09 &0.07 &0.29 &0.30 &\textbf{0.32} &0.26 &0.26 &\textbf{0.26} &0.31 &0.36 &\textbf{0.40} &0.39 &0.46 &\textbf{0.51} \\
15 &0.34 &0.03 &0.42 &\textbf{0.42} &0.42 &0.41 &0.41 &\textbf{0.41} &\textbf{0.43} &0.43 &0.43 &0.40 &0.40 &\textbf{0.41} \\
16 &0.26 &0.11 &\textbf{0.59} &0.58 &0.59 &0.57 &\textbf{0.57} &0.57 &\textbf{0.57} &0.55 &0.56 &0.63 &0.63 &\textbf{0.65} \\
17 &0.54 &0.09 &0.60 &0.60 &\textbf{0.61} &0.59 &0.59 &\textbf{0.59} &0.55 &0.55 &\textbf{0.56} &0.66 &0.65 &\textbf{0.68} \\
18 &0.52 &0.07 &0.61 &0.62 &\textbf{0.63} &0.57 &0.57 &\textbf{0.6} &0.44 &0.44 &\textbf{0.45} &0.65 &0.65 &\textbf{0.66} \\
19 &0.59 &0.06 &0.62 &0.62 &\textbf{0.65} &0.59 &0.6 &\textbf{0.62} &0.51 &0.51 &\textbf{0.53} &0.67 &0.68 &\textbf{0.71} \\
20 &0.52 &0.07 &0.58 &\textbf{0.58} &0.57 &0.54 &0.55 &\textbf{0.55} &0.47 &\textbf{0.48} &0.48 &0.58 &0.58 &\textbf{0.58} \\
21 &0.48 &0.06 &0.49 &\textbf{0.50} &0.48 &0.44 &\textbf{0.44} &0.44 &\textbf{0.39} &0.38 &0.37 &0.55 &\textbf{0.58} &0.54 \\
22 &0.39 &0.04 &0.47 &0.47 &\textbf{0.48} &0.42 &0.42 &\textbf{0.43} &0.39 &0.39 &\textbf{0.39} &0.56 &0.56 &\textbf{0.57} \\
23 &0.34 &0.04 &0.39 &0.39 &\textbf{0.39} &0.35 &\textbf{0.35} &0.35 &0.34 &\textbf{0.35} &0.34 &0.39 &0.40 &\textbf{0.40} \\
24 &0.47 &0.03 &0.50 &0.50 &\textbf{0.50} &0.46 &0.46 &\textbf{0.46} &\textbf{0.38} &0.38 &0.38 &0.55 &\textbf{0.56} &0.54 \\

\hline
COUNT &-&- &5&8&11 &2&10&12 &9&7&8 &4&9&11\\

\hline\hline
\end{tabular}\label{select_kappa}}
\end{table*}

\begin{table*}
\setlength{\abovecaptionskip}{0pt}%
\setlength{\belowcaptionskip}{10pt}%
\caption{NMI of the weighted clustering ensemble methods with different evaluation methods. For each comparison, the best result is highlighted in boldface. The last line displays the sum of the number of highlighted values in each column, which is denoted as COUNT. Note that the statistics of base partitions are presented in the second and third columns.}
\scalebox{0.55}{
\renewcommand{\arraystretch}{0.9} 
\tabcolsep 2.75mm 
\begin{tabular}{ccccccccccccccc}
\hline\hline
\multirow{2}{*}{Dataset}&\multicolumn{2}{c}{KM}
                        &\multicolumn{3}{c}{WCT+KM} &\multicolumn{3}{c}{WTQ+KM}
                        &\multicolumn{3}{c}{CSPA}&\multicolumn{3}{c}{EAC-AL}\\
                        \cmidrule(r){2-3}
                        \cmidrule(r){4-6} \cmidrule(r){7-9}
                        \cmidrule(r){10-12} \cmidrule(r){13-15}
\quad                   &Avg(base) &Std(base)
                        &ENMI &SME &F        &ENMI &SME &F
                        &ENMI &SME &F        &ENMI &SME &F \\\hline
1  &0.65 &0.04 &0.73 &\textbf{0.74} &0.74 &\textbf{0.74} &0.73 &0.74 &0.84 &\textbf{0.85} &0.83 &0.77 &\textbf{0.77} &0.76 \\
2  &0.69 &0.07 &0.86 &0.85 &\textbf{0.87} &\textbf{0.85} &0.84 &0.84 &\textbf{0.79} &0.78 &0.78 &0.87 &0.88 &\textbf{0.89} \\
3  &0.57 &0.04 &\textbf{0.68} &0.62 &0.67 &\textbf{0.68} &0.65 &0.68 &0.74 &\textbf{0.74} &0.74 &\textbf{0.71} &0.71 &0.71 \\
4  &0.36 &0.03 &0.33 &\textbf{0.34} &0.34 &0.33 &\textbf{0.34} &0.33 &\textbf{0.29} &0.29 &0.28 &\textbf{0.33} &0.33 &0.33 \\
5  &0.57 &0.06 &0.67 &\textbf{0.73} &0.72 &0.70 &0.72 &\textbf{0.72} &0.47 &0.47 &\textbf{0.47} &\textbf{0.76} &0.75 &0.74 \\
6  &0.58 &0.02 &0.58 &\textbf{0.61} &0.59 &0.58 &\textbf{0.61} &0.59 &0.53 &0.52 &\textbf{0.53} &0.63 &0.63 &\textbf{0.63} \\
7  &0.62 &0.02 &0.60 &0.60 &\textbf{0.60} &0.60 &\textbf{0.60} &0.60 &0.56 &\textbf{0.56} &0.56 &0.61 &\textbf{0.61} &0.61 \\
8  &0.33 &0.04 &0.35 &0.35 &\textbf{0.36} &0.36 &0.35 &\textbf{0.37} &\textbf{0.38} &0.38 &0.37 &0.36 &0.36 &\textbf{0.38} \\
9  &0.40 &0.04 &\textbf{0.49} &0.48 &0.48 &\textbf{0.46} &0.46 &0.46 &0.44 &0.44 &\textbf{0.44} &0.48 &\textbf{0.49} &0.48 \\
10 &0.42 &0.04 &\textbf{0.59} &0.58 &0.59 &0.60 &\textbf{0.61} &0.60 &\textbf{0.46} &0.46 &0.44 &0.51 &0.51 &\textbf{0.52} \\
11 &0.73 &0.03 &\textbf{0.78} &0.75 &0.78 &\textbf{0.80} &0.77 &0.79 &0.79 &\textbf{0.79} &0.78 &\textbf{0.78} &0.77 &0.76 \\
12 &0.24 &0.02 &\textbf{0.23} &0.17 &0.22 &\textbf{0.20} &0.13 &0.16 &0.33 &0.34 &\textbf{0.34} &0.35 &\textbf{0.36} &0.34 \\
13 &0.82 &0.05 &0.84 &0.85 &\textbf{0.86} &0.86 &0.87 &\textbf{0.88} &\textbf{0.76} &0.76 &0.76 &\textbf{0.96} &0.94 &0.94 \\
14 &0.37 &0.04 &0.38 &\textbf{0.39} &0.38 &0.37 &\textbf{0.38} &0.37 &0.37 &0.39 &\textbf{0.39} &0.41 &\textbf{0.43} &0.42 \\
15 &0.70 &0.01 &0.68 &0.69 &\textbf{0.69} &0.69 &0.69 &\textbf{0.69} &0.66 &\textbf{0.67} &0.66 &0.69 &\textbf{0.70} &0.69 \\
16 &0.55 &0.03 &0.57 &0.53 &\textbf{0.60} &0.57 &0.53 &\textbf{0.57} &0.53 &0.53 &\textbf{0.54} &0.64 &\textbf{0.65} &0.63 \\
17 &0.56 &0.07 &0.65 &\textbf{0.67} &0.66 &0.66 &\textbf{0.67} &0.66 &0.61 &\textbf{0.64} &0.63 &0.67 &0.69 &\textbf{0.70} \\
18 &0.62 &0.05 &\textbf{0.73} &0.72 &0.73 &0.72 &0.71 &\textbf{0.72} &0.62 &\textbf{0.62} &0.62 &\textbf{0.74} &0.74 &0.74 \\
19 &0.63 &0.05 &0.68 &\textbf{0.69} &0.69 &0.67 &\textbf{0.68} &0.68 &0.62 &\textbf{0.63} &0.62 &0.71 &\textbf{0.72} &0.72 \\
20 &0.61 &0.04 &\textbf{0.67} &0.66 &0.67 &0.67 &0.66 &\textbf{0.67} &\textbf{0.62} &0.62 &0.62 &\textbf{0.66} &0.65 &0.64 \\
21 &0.52 &0.04 &0.52 &0.52 &\textbf{0.52} &0.51 &0.50 &\textbf{0.51} &0.47 &0.47 &\textbf{0.47} &0.55 &0.54 &\textbf{0.55} \\
22 &0.54 &0.02 &\textbf{0.62} &0.62 &0.62 &\textbf{0.61} &0.60 &0.60 &\textbf{0.57} &0.57 &0.57 &\textbf{0.64} &0.62 &0.62 \\
23 &0.31 &0.02 &\textbf{0.35} &0.34 &0.34 &\textbf{0.33} &0.33 &0.32 &0.33 &\textbf{0.33} &0.33 &0.34 &0.35 &\textbf{0.35} \\
24 &0.58 &0.02 &0.59 &0.59 &\textbf{0.60} &\textbf{0.59} &0.59 &0.59 &0.54 &\textbf{0.55} &0.55 &0.61 &\textbf{0.61} &0.61 \\

\hline
COUNT &-&- &9&7&8 &9&7&8 &7&10&7 &8&9&7\\

\hline\hline
\end{tabular}\label{weight_nmi}}
\end{table*}

\begin{table*}
\setlength{\abovecaptionskip}{1pt}%
\setlength{\belowcaptionskip}{10pt}%
\caption{Kappa of the weighted clustering ensemble methods with different evaluation methods. For each comparison, the best result is highlighted in boldface. The last line displays the sum of the number of highlighted values in each column, which is denoted as COUNT. Note that the statistics of base partitions are presented in the second and third columns.}
\scalebox{0.55}{
\renewcommand{\arraystretch}{0.9} 
\tabcolsep 2.75mm 
\begin{tabular}{ccccccccccccccc}
\hline\hline
\multirow{2}{*}{Dataset}&\multicolumn{2}{c}{KM}
                        &\multicolumn{3}{c}{WCT+KM} &\multicolumn{3}{c}{WTQ+KM}
                        &\multicolumn{3}{c}{CSPA}&\multicolumn{3}{c}{EAC-AL}\\
                        \cmidrule(r){2-3}
                        \cmidrule(r){4-6} \cmidrule(r){7-9}
                        \cmidrule(r){10-12} \cmidrule(r){13-15}

\quad                   &Avg(base) &Std(base)
                        &ENMI &SME &F        &ENMI &SME &F
                        &ENMI &SME &F        &ENMI &SME &F\\\hline
1  &0.42 &0.12 &0.73 &0.74 &\textbf{0.74} &0.70 &0.69 &\textbf{0.71} &0.92 &\textbf{0.93} &0.91 &0.80 &\textbf{0.80} &0.80 \\
2  &0.54 &0.17 &0.91 &0.91 &\textbf{0.92} &\textbf{0.91} &0.90 &0.89 &\textbf{0.89} &0.89 &0.89 &0.95 &0.95 &\textbf{0.96} \\
3  &0.36 &0.14 &\textbf{0.80} &0.68 &0.78 &\textbf{0.78} &0.70 &0.77 &0.89 &\textbf{0.89} &0.89 &0.84 &0.84 &\textbf{0.85} \\
4  &0.20 &0.05 &0.26 &\textbf{0.27} &0.27 &0.27 &\textbf{0.27} &0.27 &\textbf{0.32} &0.31 &0.31 &\textbf{0.30} &0.28 &0.28 \\
5  &0.39 &0.20 &0.72 &\textbf{0.83} &0.83 &0.76 &0.82 &\textbf{0.82} &0.55 &0.62 &\textbf{0.63} &\textbf{0.90} &0.90 &0.89 \\
6  &0.38 &0.06 &0.45 &\textbf{0.51} &0.46 &0.44 &\textbf{0.52} &0.47 &0.42 &0.42 &\textbf{0.43} &0.55 &0.55 &\textbf{0.56} \\
7  &0.33 &0.04 &0.40 &\textbf{0.40} &0.40 &0.40 &\textbf{0.40} &0.40 &0.41 &0.42 &\textbf{0.43} &0.40 &0.41 &\textbf{0.41} \\
8  &0.20 &0.07 &0.35 &0.34 &\textbf{0.36} &0.35 &0.34 &\textbf{0.36} &\textbf{0.41} &0.40 &0.39 &0.39 &0.38 &\textbf{0.39} \\
9  &0.29 &0.13 &\textbf{0.76} &0.76 &0.75 &\textbf{0.72} &0.72 &0.72 &0.70 &0.70 &\textbf{0.70} &0.74 &\textbf{0.74} &0.74 \\
10 &0.21 &0.14 &\textbf{0.83} &0.80 &0.83 &0.83 &\textbf{0.84} &0.83 &\textbf{0.70} &0.70 &0.69 &0.77 &\textbf{0.77} &0.77 \\
11 &0.49 &0.12 &\textbf{0.69} &0.61 &0.65 &\textbf{0.67} &0.61 &0.64 &0.80 &\textbf{0.81} &0.79 &0.57 &0.62 &\textbf{0.63} \\
12 &0.16 &0.09 &\textbf{0.46} &0.36 &0.44 &\textbf{0.41} &0.29 &0.36 &0.62 &0.63 &\textbf{0.64} &0.64 &\textbf{0.65} &0.62 \\
13 &0.49 &0.14 &0.61 &0.67 &\textbf{0.68} &0.65 &\textbf{0.72} &0.71 &0.55 &0.55 &\textbf{0.56} &\textbf{0.91} &0.85 &0.85 \\
14 &0.09 &0.07 &0.31 &\textbf{0.43} &0.34 &0.27 &\textbf{0.30} &0.27 &0.35 &0.37 &\textbf{0.38} &0.45 &\textbf{0.52} &0.51 \\
15 &0.34 &0.03 &0.40 &0.40 &\textbf{0.42} &0.41 &0.40 &\textbf{0.41} &0.43 &\textbf{0.44} &0.43 &0.40 &\textbf{0.41} &0.41 \\
16 &0.26 &0.11 &0.53 &0.48 &\textbf{0.58} &0.54 &0.50 &\textbf{0.56} &0.56 &\textbf{0.59} &0.59 &0.64 &\textbf{0.65} &0.63 \\
17 &0.54 &0.09 &0.59 &0.61 &\textbf{0.61} &0.59 &0.59 &\textbf{0.60} &0.55 &\textbf{0.58} &0.57 &0.69 &0.73 &\textbf{0.73} \\
18 &0.52 &0.07 &0.64 &0.63 &\textbf{0.64} &0.60 &0.59 &\textbf{0.60} &0.44 &0.45 &\textbf{0.45} &\textbf{0.67} &0.67 &0.66 \\
19 &0.59 &0.06 &0.63 &\textbf{0.65} &0.65 &0.59 &\textbf{0.62} &0.62 &0.54 &\textbf{0.54} &0.54 &0.70 &\textbf{0.72} &0.72 \\
20 &0.52 &0.07 &\textbf{0.58} &0.57 &0.58 &0.55 &0.54 &\textbf{0.55} &\textbf{0.48} &0.46 &0.47 &0.57 &\textbf{0.57} &0.57 \\
21 &0.48 &0.06 &0.47 &0.47 &\textbf{0.48} &0.42 &0.41 &\textbf{0.43} &0.36 &0.36 &\textbf{0.36} &0.54 &0.55 &\textbf{0.55} \\
22 &0.39 &0.04 &0.48 &\textbf{0.50} &0.48 &0.43 &\textbf{0.44} &0.43 &0.39 &\textbf{0.40} &0.40 &\textbf{0.57} &0.57 &0.56 \\
23 &0.34 &0.04 &\textbf{0.40} &0.39 &0.39 &\textbf{0.35} &0.34 &0.34 &0.34 &\textbf{0.35} &0.35 &0.40 &0.41 &\textbf{0.41} \\
24 &0.47 &0.03 &0.49 &0.49 &\textbf{0.49} &\textbf{0.46} &0.46 &0.46 &0.38 &\textbf{0.38} &0.38 &0.56 &0.56 &\textbf{0.57} \\
\hline
COUNT &-&- &7&7&10 &7&8&9 &5&10&9 &5&9&10\\
\hline\hline
\end{tabular}\label{weight_kappa}}
\end{table*}

\begin{table*}
\setlength{\abovecaptionskip}{0pt}%
\setlength{\belowcaptionskip}{10pt}%
\caption{NMI of the five clustering ensemble methods. "-" indicates that the method is so time-consuming that it is hard to practice on the corresponding dataset. For each dataset, the best result is highlighted in boldface. The last line displays the average rank value for each method. Note that the statistics of base partitions are presented in the second and third columns.}
\scalebox{0.55}{
\renewcommand{\arraystretch}{0.90} 
\tabcolsep 7.5mm 
\begin{tabular}{cccccccc}
\hline\hline
\multirow{2}{*}{Dataset}&\multicolumn{2}{c}{KM}
                        &\multicolumn{3}{c}{Standard clustering ensemble}
                        &\multicolumn{2}{c}{Selective clustering ensemble} \\
                        \cmidrule(r){2-3}
                        \cmidrule(r){4-6} \cmidrule(r){7-8}
                        \quad
                        &Avg(base) &Std(base)
                        &CESHL &SPCE &TRCE
                        &DSME &DSKF
                        \\\hline
1  &  0.65 &  0.04 &           0.77 &           0.76 &           0.76 &  \textbf{0.79} &           0.77 \\
2  &  0.69 &  0.07 &  \textbf{0.89} &           0.89 &           0.88 &           0.84 &           0.87 \\
3  &  0.57 &  0.04 &           0.69 &           0.64 &           0.63 &  \textbf{0.72} &           0.71 \\
4  &  0.36 &  0.03 &           0.34 &  \textbf{0.46} &           0.32 &           0.34 &           0.35 \\
5  &  0.57 &  0.06 &           0.76 &  \textbf{0.76} &           0.75 &           0.75 &           0.73 \\
6  &  0.58 &  0.02 &           0.65 &  \textbf{0.71} &           0.63 &           0.62 &           0.63 \\
7  &  0.62 &  0.02 &           0.62 &  \textbf{0.68} &           0.65 &           0.62 &           0.61 \\
8  &  0.33 &  0.04 &           0.39 &  \textbf{0.43} &           0.34 &           0.39 &           0.38 \\
9  &  0.40 &  0.04 &           0.49 &           0.48 &           0.46 &           0.44 &   \textbf{0.50} \\
10 &  0.42 &  0.04 &  \textbf{0.66} &           0.40 &           0.56 &           0.56 &           0.56 \\
11 &  0.73 &  0.03 &           0.79 &  \textbf{0.81} &           0.80 &           0.77 &           0.76 \\
12 &  0.24 &  0.02 &  \textbf{0.39} &           0.37 &           0.35 &           0.35 &           0.32 \\
13 &  0.82 &  0.05 &              - &              - &              - &           0.94 &  \textbf{0.96} \\
14 &  0.37 &  0.04 &              - &              - &              - &  \textbf{0.41} &           0.41 \\
15 &  0.70 &  0.01 &              - &              - &              - &   \textbf{0.70} &           0.69 \\
16 &  0.55 &  0.03 &              - &              - &              - &  \textbf{0.64} &           0.63 \\
17 &  0.56 &  0.07 &           0.56 &  \textbf{0.69} &           0.67 &           0.65 &           0.66 \\
18 &  0.62 &  0.05 &           0.63 &           0.69 &  \textbf{0.75} &           0.73 &           0.74 \\
19 &  0.63 &  0.05 &           0.62 &  \textbf{0.74} &           0.69 &           0.70 &           0.69 \\
20 &  0.61 &  0.04 &           0.62 &           0.69 &  \textbf{0.69} &           0.65 &           0.65 \\
21 &  0.52 &  0.04 &           0.54 &  \textbf{0.61} &           0.55 &           0.58 &           0.57 \\
22 &  0.54 &  0.02 &              - &              - &              - &           0.62 &   \textbf{0.62} \\
23 &  0.31 &  0.02 &              - &              - &              - &           0.35 &   \textbf{0.35} \\
24 &  0.58 &  0.02 &              - &              - &              - &  \textbf{0.61} &           0.61 \\

\hline
Avg(rank) &-&- &3.06&2.06&3.35 &2.71&2.79\\

\hline\hline
\end{tabular}\label{ce_nmi}}
\end{table*}

\begin{table*}
\setlength{\abovecaptionskip}{1pt}%
\setlength{\belowcaptionskip}{10pt}%
\caption{Kappa of the five clustering ensemble methods. "-" indicates that the method is so time-consuming that it is hard to practice on the corresponding dataset. For each dataset, the best result is highlighted in boldface. The last line displays the average rank value for each method. Note that the statistics of base partitions are presented in the second and third columns.}
\scalebox{0.55}{
\renewcommand{\arraystretch}{0.9} 
\tabcolsep 7.5mm 
\begin{tabular}{cccccccc}
\hline\hline
\multirow{2}{*}{Dataset}&\multicolumn{2}{c}{KM}
                        &\multicolumn{3}{c}{Standard clustering ensemble}
                        &\multicolumn{2}{c}{Selective clustering ensemble} \\
                        \cmidrule(r){2-3}
                        \cmidrule(r){4-6} \cmidrule(r){7-8}
                        \quad
                        &Avg(base) &Std(base)
                        &CESHL &SPCE &TRCE
                        &DSME &DSKF
                        \\\hline
1  &  0.42 &  0.12 &           0.80 &           0.79 &           0.80 &  \textbf{0.84} &           0.83 \\
2  &  0.54 &  0.17 &  \textbf{0.96} &           0.96 &           0.96 &           0.94 &           0.95 \\
3  &  0.36 &  0.14 &           0.82 &           0.71 &           0.73 &  \textbf{0.85} &           0.85 \\
4  &  0.20 &  0.05 &           0.28 &           0.28 &           0.26 &           0.30 &  \textbf{0.31} \\
5  &  0.39 &  0.20 &           0.90 &  \textbf{0.90} &           0.90 &           0.89 &           0.88 \\
6  &  0.38 &  0.06 &           0.62 &  \textbf{0.75} &           0.60 &           0.52 &           0.53 \\
7  &  0.33 &  0.04 &  \textbf{0.43} &           0.40 &           0.42 &           0.41 &           0.42 \\
8  &  0.20 &  0.07 &  \textbf{0.41} &           0.36 &           0.37 &           0.40 &           0.39 \\
9  &  0.29 &  0.13 &           0.75 &           0.73 &           0.72 &           0.70 &  \textbf{0.76} \\
10 &  0.21 &  0.14 &  \textbf{0.86} &           0.55 &           0.74 &           0.80 &           0.81 \\
11 &  0.49 &  0.12 &           0.62 &           0.61 &           0.60 &  \textbf{0.67} &           0.61 \\
12 &  0.16 &  0.09 &  \textbf{0.68} &           0.64 &           0.61 &           0.64 &           0.61 \\
13 &  0.49 &  0.14 &              - &              - &              - &           0.84 &  \textbf{0.89} \\
14 &  0.09 &  0.07 &              - &              - &              - &  \textbf{0.49} &           0.48 \\
15 &  0.34 &  0.03 &              - &              - &              - &  \textbf{0.41} &           0.40 \\
16 &  0.26 &  0.11 &              - &              - &              - &  \textbf{0.64} &           0.63 \\
17 &  0.54 &  0.09 &           0.41 &           0.58 &           0.63 &           0.66 &  \textbf{0.67} \\
18 &  0.52 &  0.07 &           0.53 &           0.58 &           0.65 &           0.68 &  \textbf{0.68} \\
19 &  0.59 &  0.06 &           0.57 &           0.66 &           0.66 &  \textbf{0.70} &           0.69 \\
20 &  0.52 &  0.07 &           0.53 &           0.51 &  \textbf{0.59} &           0.57 &           0.57 \\
21 &  0.48 &  0.06 &           0.52 &           0.51 &           0.51 &  \textbf{0.59} &           0.57 \\
22 &  0.39 &  0.04 &              - &              - &              - &           0.56 &  \textbf{0.56} \\
23 &  0.34 &  0.04 &              - &              - &              - &           0.39 &   \textbf{0.40} \\
24 &  0.47 &  0.03 &              - &              - &              - &           0.58 &  \textbf{0.59} \\

\hline
Avg(rank) &-&- &2.59&3.65&3.59 &2.33&2.21\\

\hline\hline
\end{tabular}\label{ce_kappa}}
\end{table*}

\begin{table*}[!t]
\setlength{\abovecaptionskip}{0pt}%
\setlength{\belowcaptionskip}{10pt}%
\caption{Execution time (in seconds) of 6 evaluation methods and 2 ensemble methods on 24 datasets. For NMI, SMEP and Kappa, we calculate the time they took to select base partitions during the ensemble process respectively. For ENMI, SME and F, we calculate the time they took to weight clusters during the ensemble process respectively. For DSME and DSKF, the time for generating the final ensemble and weighting clusters is calculated respectively. For each comparison, the shortest runtime is highlighted in boldface. And it will be underlined if the ratio of the shortest runtime to the second shortest runtime was less than 0.5.}
\scalebox{0.55}{
\renewcommand{\arraystretch}{1.125} 
\tabcolsep 6.5mm 
\label{tabel:viii}
\begin{tabular}{ccccccccc}
\hline\hline
\multirow{2}{*}{Dataset}&\multicolumn{3}{c}{Partition evaluation methods}
                        &\multicolumn{3}{c}{Cluster evaluation methods}
                        &\multicolumn{2}{c}{Ensemble methods}\\
                        \cmidrule(r){2-4} \cmidrule(r){5-7} \cmidrule(r){8-9}
                \quad   &NMI &SMEP &Kappa
                        & ENMI & SME & F
                        & DSME & DSKF\\\hline
1 &\uline{\textbf{8.67}} &38.47 &16.56 &65.22 &18.98 &\uline{\textbf{11.61}}            &41.33 &\uline{\textbf{18.38}}\\
2 &\uline{\textbf{5.98}} &42.23 &17.08 &75.02 &22.16 &\textbf{11.91}            &48.75 &\uline{\textbf{18.80}}\\
3 &\uline{\textbf{6.88}} &45.34 &18.62 &92.47 &24.84 &\uline{\textbf{11.86}}  &52.13 &\uline{\textbf{20.33}}\\
4 &\uline{\textbf{7.05}} &52.89 &19.34 &103.69 &28.05 &\uline{\textbf{13.66}} &59.75 &\uline{\textbf{20.88}}\\
5 &\uline{\textbf{7.20}} &54.33 &18.53 &360.5 &28.48 &\uline{\textbf{12.36}} &60.67 &\uline{\textbf{20.53}}\\
6 &\uline{\textbf{9.20}} &76.39 &23.73 &830.83 &39.02 &\textbf{20.86} &79.30 &\uline{\textbf{25.17}}\\
7 &\uline{\textbf{10.78}} &125.55 &32.80 &1475.47 &70.56 &\textbf{41.84} &128.69 &\uline{\textbf{37.19}}\\
8 &\uline{\textbf{10.66}} &75.78 &23.19 &882.61 &42.00 &\uline{\textbf{18.97}} &82.69 &\uline{\textbf{25.88}}\\
9 &\uline{\textbf{11.11}} &70.16 &23.81 &591.27 &31.66 &\uline{\textbf{15.33}} &81.55 &\uline{\textbf{26.25}}\\
10 &\uline{\textbf{13.55}} &84.36 &75.67 &1148.38 &46.47 &\uline{\textbf{23.23}} &113.88 &\textbf{78.70}\\
11 &\uline{\textbf{11.34}} &92.14 &80.02 &1360.89 &55.08 &\textbf{30.44} &102.33 &\textbf{83.30}\\
12 &\uline{\textbf{17.34}} &83.11 &84.61 &1445.0 &48.67 &\textbf{26.89} &\textbf{94.94} &97.45\\
13 &\textbf{116.39} &334.73 &172.25 &5521.23 &152.75 &\textbf{103.41} &358.70 &\textbf{206.75}\\
14 &\uline{\textbf{252.61}} &5158.75 &913.91 &11991.08 &2933.34 &\textbf{1533.31} &5790.70 &\uline{\textbf{1089.64}}\\
15 &\uline{\textbf{160.94}} &5793.08 &2296.73 &21776.33 &\textbf{3768.97} &3883.28 &6229.34 &\uline{\textbf{2567.67}}\\
16 &\uline{\textbf{202.94}} &4887.61 &1067.06 &14429.05 &3152.28 &\textbf{2462.39} &5205.95 &\uline{\textbf{1567.13}}\\
17 &\uline{\textbf{9.17}} &49.52 &20.44 &337.13 &22.03 &\uline{\textbf{9.45}} &57.67 &\uline{\textbf{23.63}}\\
18 &\textbf{12.05} &62.61 &22.66 &428.44 &27.00 &\uline{\textbf{11.41}} &76.56 &\uline{\textbf{31.64}}\\
19 &\uline{\textbf{15.55}} &77.27 &81.81 &728.59 &35.45 &\uline{\textbf{16.75}} &82.69 &\textbf{76.89}\\
20 &\uline{\textbf{18.05}} &84.05 &99.45 &850.36 &39.36 &\uline{\textbf{19.14}} &95.52 &\textbf{89.44}\\
21 &\uline{\textbf{15.83}} &63.33 &101.94 &615.22 &28.33 &\textbf{14.25} &\textbf{73.41} &89.89\\
22 &\textbf{108.95} &325.64 &134.80 &2605.56 &123.88 &\textbf{121.64} &410.31 &\uline{\textbf{180.48}}\\
23 &\textbf{104.47} &181.17 &187.69 &1107.31 &\textbf{51.80} &89.17 &194.80 &\textbf{191.84}\\
24 &\textbf{120.28} &323.19 &221.00 &3162.80 &129.66 &\textbf{129.36} &427.27 &\textbf{244.42}\\
\hline
COUNT &24&0&0 &0&2&22 &2&22\\
\hline\hline
\label{runtime}
\end{tabular}}
\end{table*}

\subsubsection{The efficiency analysis}
\label{running_time}
In addition to its effectiveness, we also compare the runtime of different methods, which is shown in Fig.\ref{run_time} and Table \ref{tabel:viii}. The software used for all the experiments is Matlab, and the machine is Intel Core i5 8th generation processor with 8GB of RAM.

From the figure and table, one can see that: 1) The three standard clustering ensemble methods are so time-consuming that they are difficult to practice on large datasets, and DSME and DSKF are more suitable for large datasets. For example, on the dataset \textit{Wap} with only 1560 samples, the TRCE runs for more than 10h, while the CESHL and SPCE exceed 24h. On the dataset \textit{Tr31} with 927 samples, the running times of CESHL and TRCE are more than 200 times that of DSME and DSKF, and the one of SPCE is more than 300 times that of DSME and DSKF. 2) In most cases, especially on large datasets, the computation time of DSKF is the shortest and is often less than half of the baseline ones.

In summary: 1) The proposed DSKF is insensitive to its parameters. 2) Compared to the three standard clustering ensemble methods, DSME and DSKF are more effective and efficient. The performance of DSKF is competitive compared to DSME. 3) The method is efficient and is suitable for large datasets.

\section{Conclusion}

\label{sect6}
In this paper, we first review the evaluation problem in clustering ensemble. In summary, 1) NMI ignores importance of small clusters, has finite size effect, violates proportionality assumption, and cannot evaluate the quality of a single cluster of interest. 2) To handle the last drawback that exists in NMI, four indices extend NMI. However,  simple extensions cannot solve the first three drawbacks that exist in NMI. In addition, they also suffer from the symmetric problem or context meaning problem. 3) In order to solve the problems in the NMI-type methods, SME was proposed. But it is computationally inefficient since it may go through every cluster in the reference partition for evaluation of a single cluster and the number of clusters that need to be evaluated is usually very large. 4) We propose a more efficient evaluation method for partitions and clusters using kappa and F-score. After that, we propose DSKF, which uses kappa to select diverse base partitions and uses F-score to give more weight to the more stable clusters.

Empirical results reveal that the proposed methods have better performances. Moreover, one can observe that NMI values are misleading and fail to accurately reflect the actual performance of the partitions. The performance analysis of unsupervised methods should be based on kappa rather than NMI. Interesting problems for future work include overlapping clustering ensemble, community detection ensemble, etc.




\bibliographystyle{elsarticle-num}
\bibliography{references.bib}

\begin{thebibliography}{10}
\expandafter\ifx\csname url\endcsname\relax
  \def\url#1{\texttt{#1}}\fi
\expandafter\ifx\csname urlprefix\endcsname\relax\def\urlprefix{URL }\fi
\expandafter\ifx\csname href\endcsname\relax
  \def\href#1#2{#2} \def\path#1{#1}\fi

\bibitem{jain1999data}
A.~K. Jain, M.~N. Murty, P.~J. Flynn, Data clustering: a review, ACM computing
  surveys (CSUR) 31~(3) (1999) 264--323.

\bibitem{boongoen2018cluster}
T.~Boongoen, N.~Iam-On, Cluster ensembles: A survey of approaches with recent
  extensions and applications, Computer Science Review 28 (2018) 1--25.

\bibitem{strehl2003cluster}
A.~Strehl, J.~Ghosh, Cluster ensembles---a knowledge reuse framework for
  combining multiple partitions, Journal of machine learning research 3~(Dec)
  (2002) 583--617.

\bibitem{nguyen2007consensus}
N.~Nguyen, R.~Caruana, Consensus clusterings, in: Seventh IEEE international
  conference on data mining (ICDM 2007), IEEE, 2007, pp. 607--612.

\bibitem{vegapons2011a}
S.~Vega-Pons, J.~Ruiz-Shulcloper, A survey of clustering ensemble algorithms,
  International Journal of Pattern Recognition and Artificial Intelligence
  25~(03) (2011) 337--372.

\bibitem{wu2018a}
X.~Wu, T.~Ma, J.~Cao, Y.~Tian, A.~Alabdulkarim, A comparative study of
  clustering ensemble algorithms, Computers \& Electrical Engineering 68 (2018)
  603--615.

\bibitem{zhang2019weighted}
M.~Zhang, Weighted clustering ensemble: A review, arXiv preprint
  arXiv:1910.02433 (2019).

\bibitem{tsai2012cluster}
C.-F. Tsai, C.~Hung, Cluster ensembles in collaborative filtering
  recommendation, Applied Soft Computing 12~(4) (2012) 1417--1425.

\bibitem{zheng2013penetrate}
L.~Zheng, L.~Li, W.~Hong, T.~Li, Penetrate: Personalized news recommendation
  using ensemble hierarchical clustering, Expert Systems with Applications
  40~(6) (2013) 2127--2136.

\bibitem{logesh2020enhancing}
R.~Logesh, V.~Subramaniyaswamy, D.~Malathi, N.~Sivaramakrishnan,
  V.~Vijayakumar, Enhancing recommendation stability of collaborative filtering
  recommender system through bio-inspired clustering ensemble method, Neural
  Computing and Applications 32~(7) (2020) 2141--2164.

\bibitem{wang2014breast}
C.~Wang, R.~Machiraju, K.~Huang, Breast cancer patient stratification using a
  molecular regularized consensus clustering method, Methods 67~(3) (2014)
  304--312.

\bibitem{liu2017entropy}
H.~Liu, R.~Zhao, H.~Fang, F.~Cheng, Y.~Fu, Y.-Y. Liu, Entropy-based consensus
  clustering for patient stratification, Bioinformatics 33~(17) (2017)
  2691--2698.

\bibitem{zhang2019a}
Y.-Y. Zhang, C.~Yang, J.~Wang, C.-H. Zheng, A link and weight-based ensemble
  clustering for patient stratification, in: International Conference on
  Intelligent Computing, Springer, 2019, pp. 256--264.

\bibitem{zhang2008spectral}
X.~Zhang, L.~Jiao, F.~Liu, L.~Bo, M.~Gong, Spectral clustering ensemble applied
  to sar image segmentation, IEEE Transactions on Geoscience and Remote Sensing
  46~(7) (2008) 2126--2136.

\bibitem{kuo2016an}
R.-J. Kuo, C.~Mei, F.~E. Zulvia, C.~Tsai, An application of a metaheuristic
  algorithm-based clustering ensemble method to app customer segmentation,
  Neurocomputing 205 (2016) 116--129.

\bibitem{shi2018transfer}
Y.~Shi, Z.~Yu, C.~P. Chen, J.~You, H.-S. Wong, Y.~Wang, J.~Zhang, Transfer
  clustering ensemble selection, IEEE transactions on cybernetics 50~(6) (2018)
  2872--2885.

\bibitem{li2008weighted}
T.~Li, C.~Ding, Weighted consensus clustering, in: Proceedings of the 2008 SIAM
  International Conference on Data Mining, SIAM, 2008, pp. 798--809.

\bibitem{hadjitodorov2006moderate}
S.~T. Hadjitodorov, L.~I. Kuncheva, L.~P. Todorova, Moderate diversity for
  better cluster ensembles, Information Fusion 7~(3) (2006) 264--275.

\bibitem{fern2008cluster}
X.~Z. Fern, W.~Lin, Cluster ensemble selection, Statistical Analysis and Data
  Mining: The ASA Data Science Journal 1~(3) (2008) 128--141.

\bibitem{azimi2009adaptive}
J.~Azimi, X.~Z. Fern, Adaptive cluster ensemble selection., in: Ijcai, Vol.~9,
  2009, pp. 992--997.

\bibitem{hong2009resampling}
Y.~Hong, S.~Kwong, H.~Wang, Q.~Ren, Resampling-based selective clustering
  ensembles, Pattern recognition letters 30~(3) (2009) 298--305.

\bibitem{jia2011bagging}
J.~Jia, X.~Xiao, B.~Liu, L.~Jiao, Bagging-based spectral clustering ensemble
  selection, Pattern Recognition Letters 32~(10) (2011) 1456--1467.

\bibitem{alizadeh2011new}
H.~Alizadeh, B.~Minaei, H.~Parvin, A new criterion for clusters validation, in:
  Artificial Intelligence Applications and Innovations, Springer, 2011, pp.
  110--115.

\bibitem{alizadeh2011asymmetric}
H.~Alizadeh, B.~Minaei, H.~Parvin, M.~Moshki, An asymmetric criterion for
  cluster validation, in: Developing Concepts in Applied Intelligence,
  Springer, 2011, pp. 1--14.

\bibitem{li2018cluster}
F.~Li, Y.~Qian, J.~Wang, C.~Dang, B.~Liu, Cluster’s quality evaluation and
  selective clustering ensemble, ACM Transactions on Knowledge Discovery from
  Data (TKDD) 12~(5) (2018) 1--27.

\bibitem{abbasi2019clustering}
S.-o. Abbasi, S.~Nejatian, H.~Parvin, V.~Rezaie, K.~Bagherifard, Clustering
  ensemble selection considering quality and diversity, Artificial Intelligence
  Review 52~(2) (2019) 1311--1340.

\bibitem{naldi2013cluster}
M.~C. Naldi, A.~Carvalho, R.~J. Campello, Cluster ensemble selection based on
  relative validity indexes, Data Mining and Knowledge Discovery 27~(2) (2013)
  259--289.

\bibitem{zhou2006clusterer}
Z.-H. Zhou, W.~Tang, Clusterer ensemble, Knowledge-Based Systems 19~(1) (2006)
  77--83.

\bibitem{gullo2009diversity}
F.~Gullo, A.~Tagarelli, S.~Greco, Diversity-based weighting schemes for
  clustering ensembles, in: Proceedings of the 2009 SIAM international
  conference on data mining, SIAM, 2009, pp. 437--448.

\bibitem{alhichri2014clustering}
H.~Alhichri, N.~Ammour, N.~Alajlan, Y.~Bazi, Clustering of hyperspectral images
  with an ensemble method based on fuzzy c-means and markov random fields,
  Arabian Journal for Science and Engineering 39~(5) (2014) 3747--3757.

\bibitem{berikov2017ensemble}
V.~Berikov, I.~Pestunov, Ensemble clustering based on weighted co-association
  matrices: Error bound and convergence properties, Pattern Recognition 63
  (2017) 427--436.

\bibitem{yang2016overlapping}
L.~Yang, Z.~Yu, J.~Qian, S.~Liu, Overlapping community detection using weighted
  consensus clustering, Pramana 87~(4) (2016) 1--6.

\bibitem{yousefnezhad2018woce}
M.~Yousefnezhad, S.-J. Huang, D.~Zhang, Woce: A framework for clustering
  ensemble by exploiting the wisdom of crowds theory, IEEE transactions on
  cybernetics 48~(2) (2017) 486--499.

\bibitem{unlu2019a}
R.~{\"U}nl{\"u}, P.~Xanthopoulos, A weighted framework for unsupervised
  ensemble learning based on internal quality measures, Annals of Operations
  Research 276~(1) (2019) 229--247.

\bibitem{law2004multiobjective}
M.~H. Law, A.~P. Topchy, A.~K. Jain, Multiobjective data clustering, in:
  Proceedings of the 2004 IEEE Computer Society Conference on Computer Vision
  and Pattern Recognition, 2004. CVPR 2004., Vol.~2, IEEE, 2004, pp. II--II.

\bibitem{liu2019evaluation}
X.~Liu, H.-M. Cheng, Z.-Y. Zhang, Evaluation of community detection methods,
  IEEE Transactions on Knowledge and Data Engineering 32~(9) (2019) 1736--1746.

\bibitem{monti2003consensus}
S.~Monti, P.~Tamayo, J.~Mesirov, T.~Golub, Consensus clustering: a
  resampling-based method for class discovery and visualization of gene
  expression microarray data, Machine learning 52~(1) (2003) 91--118.

\bibitem{zhang2015evaluating}
P.~Zhang, Evaluating accuracy of community detection using the relative
  normalized mutual information, Journal of Statistical Mechanics: Theory and
  Experiment 2015~(11) (2015) P11006.

\bibitem{Lai2016A}
D.~Lai, C.~Nardini, A corrected normalized mutual information for performance
  evaluation of community detection, Journal of Statistical Mechanics: Theory
  and Experiment 2016~(9) (2016) 093403.

\bibitem{powers2011evaluation}
D.~M. Powers, Evaluation: from precision, recall and f-measure to roc,
  informedness, markedness and correlation, arXiv preprint arXiv:2010.16061
  (2020).

\bibitem{metz1978basic}
C.~E. Metz, Basic principles of roc analysis, in: Seminars in nuclear medicine,
  Vol.~8, Elsevier, 1978, pp. 283--298.

\bibitem{galton1892}
F.~Galton, Finger prints, no. 57490-57492, Macmillan and Company, 1892.

\bibitem{macqueen1967some}
J.~MacQueen, et~al., Some methods for classification and analysis of
  multivariate observations, in: Proceedings of the fifth Berkeley symposium on
  mathematical statistics and probability, Vol.~1, Oakland, CA, USA, 1967, pp.
  281--297.

\bibitem{iamon2011a}
N.~Iam-On, T.~Boongoen, S.~Garrett, C.~Price, A link-based approach to the
  cluster ensemble problem, IEEE transactions on pattern analysis and machine
  intelligence 33~(12) (2011) 2396--2409.

\bibitem{johnson1967hierarchical}
S.~C. Johnson, Hierarchical clustering schemes, Psychometrika 32~(3) (1967)
  241--254.

\bibitem{steinbach2000a}
M.~Steinbach, G.~Karypis, V.~Kumar, A comparison of document clustering
  techniques (2000).

\bibitem{kuncheva2004using}
L.~I. Kuncheva, S.~T. Hadjitodorov, Using diversity in cluster ensembles, in:
  2004 IEEE International Conference on Systems, Man and Cybernetics (IEEE Cat.
  No. 04CH37583), Vol.~2, IEEE, 2004, pp. 1214--1219.

\bibitem{xu2016an}
S.~Xu, K.-S. Chan, J.~Gao, X.~Xu, X.~Li, X.~Hua, J.~An, An integrated
  k-means--laplacian cluster ensemble approach for document datasets,
  Neurocomputing 214 (2016) 495--507.

\bibitem{ayad2010on}
H.~G. Ayad, M.~S. Kamel, On voting-based consensus of cluster ensembles,
  Pattern Recognition 43~(5) (2010) 1943--1953.

\bibitem{fred2005combining}
A.~L. Fred, A.~K. Jain, Combining multiple clusterings using evidence
  accumulation, IEEE transactions on pattern analysis and machine intelligence
  27~(6) (2005) 835--850.

\bibitem{zhou2022clustering}
P.~Zhou, X.~Wang, L.~Du, X.~Li, Clustering ensemble via structured hypergraph
  learning, Information Fusion 78 (2022) 171--179.

\bibitem{zhou2020self}
P.~Zhou, L.~Du, X.~Liu, Y.-D. Shen, M.~Fan, X.~Li, Self-paced clustering
  ensemble, IEEE transactions on neural networks and learning systems 32~(4)
  (2020) 1497--1511.

\bibitem{zhou2021tri}
P.~Zhou, L.~Du, Y.-D. Shen, X.~Li, Tri-level robust clustering ensemble with
  multiple graph learning, in: Thirty-Fifth AAAI Conference on Artificial
  Intelligence, 2021, pp. 11125--11133.

\end{thebibliography}



\end{document}